\documentclass[english]{article}
\usepackage[T1]{fontenc}
\usepackage[latin9]{inputenc}
\usepackage{geometry}
\usepackage[unicode=true,
 bookmarks=false,
 breaklinks=false,
 pdfborder={0 0 1},
 colorlinks=false]{hyperref}
\hypersetup{colorlinks,
linkcolor=red,
anchorcolor=blue,
citecolor=blue,
urlcolor=blue}
\usepackage{bm}
\usepackage{amsmath}
\usepackage{mathtools}
\usepackage{amssymb}
\usepackage{amsthm}
\usepackage{comment}
\usepackage{natbib}
\usepackage{booktabs}
\usepackage{graphicx}
\usepackage{algorithm}
\usepackage{algpseudocode}
\usepackage{float}
\usepackage{multirow}
\usepackage{dsfont}
\usepackage{tcolorbox}
\usepackage{tabulary}
\usepackage{colortbl}
\usepackage{color}
\usepackage[outdir=./]{epstopdf}
\usepackage{enumitem}

\definecolor{gen}{RGB}{0,0,200}
\definecolor{cc}{RGB}{231,117,0}

\allowdisplaybreaks

\newcommand{\ind}{\mathds{1}}
\newcommand{\defn}{\coloneqq}

\newcommand{\cB}{\mathcal{B}}

\newcommand{\cE}{\mathcal{E}}

\newcommand{\cN}{\mathcal{N}}

\newcommand{\bP}{\mathbb{P}}
\newcommand{\bR}{\mathbb{R}}

\newcommand{\diff}{\,\mathrm{d}}

\newcommand{\numpf}[2]{\overset{(\mathrm{#1})}{#2}}
\newcommand{\ol}{\overline}
\newcommand{\wh}{\widehat}
\newcommand{\wt}{\widetilde}

\newcommand{\data}{\mathsf{data}}
\newcommand{\sde}{\mathsf{sde}}

\newcommand{\tr}{\mathsf{tr}}
\newcommand{\veps}{\varepsilon}

\newcommand{\setc}{\mathrm{c}}

\newcommand{\score}{\mathsf{sc}}
\newcommand{\disteq}{\overset{\mathrm d}=}

\newcommand{\mymid}{\mid}

\newcommand{\frob}{\mathsf{F}}

\newcommand{\KL}{\mathsf{KL}}
\newcommand{\TV}{\mathsf{TV}}

\title{Diffusion Models Adapt to Low-Dimensional Structure Under Flexible Coefficient Choices\footnotetext{Corresponding author: Gen Li.}}

\author{Changxiao Cai\footnote{The authors contributed equally.}~\thanks{Department of Industrial and Operations Engineering, University of Michigan, Ann Arbor, USA; Email: \href{mailto:cxcai@umich.edu}{cxcai@umich.edu}.}
\and
Yuchen Jiao\footnotemark[1] \thanks{Department of Statistics and Data Science, Chinese University of Hong Kong, Hong Kong; Email: \href{mailto:yuchenjiao@cuhk.edu.hk}{\{yuchenjiao},\href{mailto:genli@cuhk.edu.hk}{genli\}@cuhk.edu.hk}.}
\and
Gen Li\footnotemark[3]
}
\date{\today}

\begin{document}

\theoremstyle{plain} 
\newtheorem{lemma}{\bf Lemma} 
\newtheorem{proposition}{\bf Proposition}
\newtheorem{theorem}{\bf Theorem}
\newtheorem{corollary}{\bf Corollary} 
\newtheorem{claim}{\bf Claim}

\theoremstyle{remark}
\newtheorem{assumption}{\bf Assumption} 
\newtheorem{definition}{\bf Definition} 
\newtheorem{condition}{\bf Condition}
\newtheorem{property}{\bf Property} 
\newtheorem{example}{\bf Example}
\newtheorem{fact}{\bf Fact}
\newtheorem{remark}{\bf Remark}

\newcommand{\figpath}{./}

\maketitle 

\begin{abstract}

Diffusion models are known to exploit unknown low-dimensional structure to accelerate sampling. However, existing convergence theory under low-dimensional data structure has largely focused on update rules with narrowly prescribed coefficient choices. This raises a fundamental question: is adaptation to low-dimensional structure sensitive to the precise choice of update coefficients? In this paper, we show that such adaptation is a robust property of diffusion models. For a broad class of
update coefficients, we prove that \(\widetilde{O}(k/\varepsilon)\) iterations suffice to generate an \(\varepsilon\)-accurate sample in total variation (TV) distance, independently of the ambient dimension. Our framework substantially broadens the class of diffusion samplers known to enjoy low dimensional adaptation and applies to several commonly used methods in practice. These results provide a theoretical justification for the empirical effectiveness of diffusion samplers across different coefficient choices when applied to structured, high-dimensional data.

\end{abstract}

\medskip


\tableofcontents

\section{Introduction}
\label{sec:intro}

Diffusion models \citep{sohl2015deep,ho2020denoising,song2019generative,song2020denoising} have emerged as a powerful class of generative models, achieving remarkable performance across a wide range of high-dimensional data-generation tasks \citep{ramesh2022hierarchical,popov2021grad,dhariwal2021diffusion}. At a high level, these models generate samples by progressively transforming Gaussian noise into data through a sequence of iterative denoising updates guided by learned score functions \citep{song2020score}. Despite their empirical success, this iterative sampling procedure can be computationally expensive, motivating extensive efforts to understand when and how diffusion sampling can be accelerated. A particularly promising direction exploits the observation that high-dimensional data often exhibit much low-dimensional geometric structure \citep{pope2021intrinsic}. Recent theoretical developments have shown that diffusion models can adapt to such unknown structure, with the number of sampling iterations depending on the intrinsic dimension of the data rather than the potentially much larger ambient dimension \citep{li2024adapting,li2024d,huang2024denoising,liang2025low,potaptchik2024linear}. Understanding the generality and robustness of this low-dimensional adaptation is therefore central to developing both an accurate theory and more efficient diffusion samplers.

To set the stage for formal discussion, given pre-trained score function estimates $\{s_t(\cdot)\}_{t=1}^T$, a diffusion-based sampling procedure can be expressed by the following general form:
\begin{align}
Y_T\sim \mathcal{N}(0, I_d) \qquad \text{and} \qquad Y_{t-1} = \frac{1}{\sqrt{\alpha_{t}}}\big(Y_t+\eta_ts_{t}(Y_t) + \sigma_t Z_t\big),\quad t = T,\cdots, 1, \label{eq:DDPM}
\end{align}
where $Z_t \overset{\mathsf{i.i.d.}}{\sim} \cN(0,I_d)$ is independent Gaussian noise, $(\alpha_t)_{t=1}^T$ is a noise schedule, and $(\eta_t, \sigma_t^2)_{t=1}^T$ are properly chosen update coefficients.

A central design choice in diffusion samplers lies in the selection of the update coefficients $(\eta_t, \sigma_t^2)$.
As the update rules of diffusion models can be equivalently interpreted as numerical discretizations of a reverse-time stochastic differential equation (SDE), these coefficients are typically chosen to match the drift and diffusion coefficients of the SDE.
Different discretization schemes therefore lead to different coefficients $(\eta_t, \sigma_t^2)$ and update rules.
For instance, the coefficients of the original DDPM update rule \citep{ho2020denoising} correspond to an exponential-integrator discretization of the SDE.

Despite their algorithmic importance, the theoretical role of these coefficients remains inadequately understood.
One prominent line of research focuses on its role on the sampling convergence when the data distribution exhibits low-dimensional structure \citep{li2024adapting,potaptchik2024linear,huang2024denoising,liang2025low,li2024d,tang2026adaptivity}.
It investigates whether the iteration complexity to achieve an desired accuracy depends on the intrinsic dimension, instead of the ambient dimension, of the target data distribution.
%
Existing convergence analyses under low-dimensional structure have largely relied on carefully calibrated update coefficients tied to specific discretization schemes. This has naturally led to the prevailing view that such calibration may be essential for exploiting unknown low-dimensional data geometry, and that even modest coefficient perturbations could substantially degrade sampling performance. 
This viewpoint, however, is in tension with empirical practice: diffusion samplers with noticeably different coefficient choices often perform well in practical applications.
These observations suggest that successful adaptation to low-dimensional structure may not rely on a narrowly prescribed choice of drift and variance coefficients.

\begin{figure}[tbp]
    \centering
    \includegraphics[width=0.5\linewidth]{\figpath/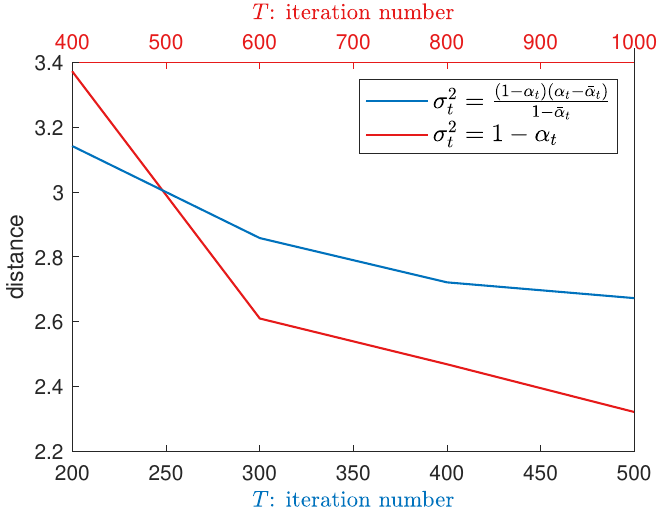}
    \caption{Comparison of two parameter choices on CIFAR-10.}
    \label{fig:cifar10}
\end{figure}

To illustrate this point, we conduct experiments on the CIFAR-10 dataset \citep{krizhevsky2009learning} using Elucidated Diffusion Models (EDM) \citep{karras2022elucidating}.
We compare two parameter choices of $\sigma_t$ across different numbers of sampling iterations.
The results are presented in Fig.~\ref{fig:cifar10}.
For ease of comparison, the figure uses two horizontal axes: the bottom axis shows the number of iterations for the choice $\sigma_t^2=\frac{(1-\alpha_t)(\alpha_t-\overline{\alpha}_t)}{1-\overline{\alpha}_t}$, while the top axis shows the corresponding number of iterations for the choice $\sigma_t^2 = 1-\alpha_t$. 
The top axis is a factor-2 rescaling of the bottom axis, reflecting that the latter sampler uses twice as many iterations.
After this constant-factor adjustment, the two parameter choices achieve comparable sampling performance.

This motivates the central question of our work:
\begin{center}
\textit{Is adaptation to low-dimensional data structure in diffusion models robust to coefficient choices?}
\end{center}

\paragraph{Our contribution}
In this paper, we answer this question affirmatively. We show that adaptation to low-dimensional structure is a robust property of diffusion samplers. Specifically, for a broad range of update coefficient choices, the number of iterations required to produce an $\varepsilon$-accurate sample in total variation (TV) distance is at most
\begin{align*}
\frac{k}{\veps} \quad \text{(up to logarithmic factors)},
\end{align*}
where $k$ represents the intrinsic dimension of the data, provided that an accurate pre-trained score function is available. 
%
%
Our theory substantially broadens the class of diffusion samplers known to enjoy provable guarantees under low-dimensional data structure. Previously, such guarantees were established only for a narrow range of coefficients tied to particular discretization schemes. In contrast, our framework accommodates a much wider family of samplers and impose substantially weaker conditions on the coefficient choices. In particular, our results apply to a broad range of methods commonly used in practice, including the original DDPM sampler, improved DDPM samplers, and analytic DPM-type samplers.
This provides a theoretical justification for the empirical robustness of diffusion samplers with different coefficient choices when sampling from structured data distributions.

\section{Preliminaries and problem setup}
\label{sec:problem}


\paragraph*{Forward process}
Let $X_0 \sim p_{\mathsf{data}}$ denote the target data distribution on $\bR^d$. The forward process is defined by
\begin{align}
\label{eq:forward}
X_t =\sqrt{\alpha_t}X_{t-1} + \sqrt{1-\alpha_t} Z_t, \quad t=1,2,\dots,T,
\end{align}
where $(\alpha_t)_{t=1}^T$ are the noise schedules and $Z_1,\dots,Z_T\overset{\mathrm{i.i.d.}}{\sim}\mathcal{N}(0,I_d)$ are $d$-dimensional independent standard Gaussian random vectors.
For convenience, define
\begin{align}
\label{eq:overline-alpha}
\overline{\alpha}_t:=\prod_{k=1}^t\alpha_k,\quad t=1,2,\dots,T.
\end{align}
Then the marginal distribution at time $t$ admits the representation
\begin{align}
	X_t =\sqrt{\ol \alpha_t}X_{0} + \sqrt{1-\ol\alpha_t} \,\ol Z_t,\quad t=1,2,\dots,T, \label{eq:forward-dist}
\end{align}
where $\ol Z_t\sim\mathcal{N}(0,I_d)$ is a standard Gaussian random vector independent of $X_{0}$.
%
In particular, when $\ol\alpha_T$ is sufficiently small, the terminal distribution $p_{X_T}$ is close to the standard Gaussian distribution on $\bR^d$.

In this work, we focus on the noise schedule $(\alpha_t)_{t\in[T]}$ that satisfies the following condition:
\begin{align}
\label{eq:learning-rate}
\overline{\alpha}_{T} = \frac{1}{T^{C_0}},
\qquad\text{and}\qquad
\overline{\alpha}_{t-1} = \overline{\alpha}_{t} + C_1\frac{\log T}{T} \overline{\alpha}_{t}(1-\overline{\alpha}_{t}), \quad t = T,\dots,2,
\end{align}
where $C_0, C_1 > 0$ are sufficiently large absolute constants with $C_1/C_0$ large enough. 

The continuous-time limit of \eqref{eq:forward} is given by the SDE
\begin{align}
\label{eq:forward-SDE}
\mathrm{d}X_{t} = -\frac{1}{2}\beta_t X_{t}\,\mathrm{d}t + \sqrt{\beta_t} \,\mathrm{d}B_{t}, \quad X_0 \sim p_{\data}; \quad t\in[0,T]
\end{align}
where $\beta_t:[0,T]\rightarrow \bR$ is a prescribed noise-rate function and $(B_t)_{t\in[0,T]}$ is a standard Brownian motion in $\bR^d$.

\paragraph*{Reverse process}
Classical time-reversal results for diffusion processes \citep{anderson1982reverse,haussmann1986time} show that if $(X_t)_{t\in[0,T]}$ solves the forward SDE \eqref{eq:forward-SDE}, then the reverse-time process $(Y^\sde_t)_{t\in[0,T]}$, defined by $Y^\sde_t \defn X_{T-t}$, satisfies
\begin{align}
	\label{eq:reverse-SDE}
	\mathrm{d}Y_t^\sde = \frac12\beta_{T-t}\Big( Y_t^\sde + 2\nabla \log p_{X_{T-t}}\big(Y_t^\sde\big)\Big)\diff t + \sqrt{\beta_{T-t}}\diff B_t, \quad Y_0^\sde \sim p_{X_T}; \quad t\in[0,T].
\end{align}
Here, $p_{X_t}$ denotes the marginal distribution of $X_t$ in \eqref{eq:forward-SDE}, and $\nabla\log p_{X_t}(x)$ is the \emph{score function} of $p_{X_t}$, which we denote by $s_t^\star(x)$.
A diffusion sampler seeks to generate samples from $p_{\data}$ by simulating the reverse SDE \eqref{eq:reverse-SDE} starting from $Y_0^\sde\sim p_{X_T}$, which is approximately standard Gaussian.

\paragraph{Update coefficient choices.}

Different update coefficient choices correspond to different discretizations of the underlying reverse-time dynamics and lead to different practical samplers. We summarize several representative examples below.
\begin{subequations}
\label{eq:coeff-choices}
\begin{itemize}
\item \textit{DDPM}. The original DDPM sampler \citep{ho2020denoising} uses 
\begin{align}\label{eq:coeff-choices-ho}
\eta_t = 1-\alpha_t \quad \text{and} \quad \sigma_t^2 \in \bigg\{\frac{\alpha_t-\overline{\alpha}_{t}}{1-\overline{\alpha}_t}(1-\alpha_t),1-\alpha_t \bigg\}.
\end{align}

\item \textit{Improved DDPM}. 
Improved DDPM \citep{nichol2021improved} takes the same drift coefficient as the original DDPM
but allows the variance coefficient to vary within an interval, i.e.,
\begin{align}
 \eta_t = 1-\alpha_t \quad \text{and} \quad\sigma_t^2 \in \bigg[\frac{\alpha_t-\overline{\alpha}_{t}}{1-\overline{\alpha}_t}(1-\alpha_t), 1-\alpha_t \bigg].
\end{align}
In this case, the specific value of $\sigma_t^2$ is learned from data, allowing the sampler to interpolate between the two variance choices above.

\item \textit{Analytic-DPM.} Analytic-DPM \citep{bao2022analytic} retains the drift coefficient of the original DDPM but replaces the prespecified variance with an analytically derived, score-dependent variance. Under our parametrization, its coefficients are given by
\begin{align}
\eta_t &= 1-\alpha_t, \quad \text{and} \quad
\sigma_t^2=
(1-\alpha_t)
\big(
1-(1-\alpha_t)\Gamma_t
\big),
\label{eq:coeff-choices-analytic}
\end{align}
where
\begin{align*}
\Gamma_t
\coloneqq
\frac{1}{d}
\mathbb{E}_{X_t\sim p_{X_t}}
\left[
\left\|s_t(X_t)\right\|_2^2
\right].
\end{align*}
In practice, $\Gamma_t$ is estimated using Monte Carlo samples:
\begin{align*}
\widehat{\Gamma}_t=
\frac{1}{M}
\sum_{m=1}^M
\frac{\big\|s_t(X_t^{(m)})\big\|_2^2}{d},
\qquad
X_t^{(m)}
\overset{\mathsf{i.i.d.}}{\sim}
p_{X_t}.
\end{align*}
The resulting variance estimate is further clipped using analytically derived lower and upper bounds to mitigate the effect of score estimation error. 
\footnote{With probability at least $1-\exp(-\Omega(d))$, it holds that $(1-\overline{\alpha}_t)\|s_t(X_t)\|_2^2 \lesssim d$.
Hence, in what follows, we consider $\widehat{\Gamma}_t \lesssim 1/(1-\overline{\alpha}_t)$.}

\end{itemize}

From a theoretical perspective, most existing analysis focuses on a small number of specific coefficient choices. For example, several works \citep{li2024d} study the simple choice
\begin{align}
(\eta_t,\sigma_t^2)=(1-\alpha_t, 1-\alpha_t),
\end{align}
while others \citep{chen2023improved,benton2023nearly} analyze
\begin{align}
(\eta_t,\sigma_t^2)=\big(2(1-\sqrt{\alpha_t}), 1-\alpha_t\big),
\end{align}
which arises naturally from certain discretizations of the reverse-time SDE. These choices are mathematically convenient and closely tied to particular numerical schemes.
	
\end{subequations}






Intuitively, the samplers discussed above can all be viewed as certain first-order discretizations of a continuous-time reverse-time SDE. This perspective suggests that the coefficients should satisfy $\eta_t,\sigma_t^2\asymp 1-\alpha_t$, while different samplers may only differ in higher-order terms. One may therefore expect that such higher-order differences should not fundamentally affect sampling performance. 

Motivated by this intuition, the goal of this paper is to develop a unified theory for DDPM-type samplers of the general form \eqref{eq:DDPM}. In particular, we aim to understand whether precise coefficient choices are essential for efficient sampling under low-dimensional data structure, or whether low-dimensional adaptation is robust across a broader class of discretization schemes.

\paragraph{Intrinsic low-dimensional structure.}
Let $\mathcal{X}\subseteq\mathbb{R}^{d}$ denote the support of the
target data distribution $p_{\mathsf{data}}$, namely, the smallest
closed set $C\subseteq\mathbb{R}^{d}$ satisfying $p_{\mathsf{data}}(C)=1$.
To characterize the intrinsic dimension of $\mathcal{X}$ in a general manner, we use the notions of $\varepsilon$-nets and covering numbers; see, for example, \citet{vershynin2018high}.

For any $\varepsilon>0$, a set $\mathcal{N}_{\varepsilon}\subseteq\mathcal{X}$ is called an $\varepsilon$-net of $\mathcal{X}$ if, for any $x\in\mathcal{X}$,
there exists some $x'$ in $\mathcal{N}_{\varepsilon}$ such that
$\Vert x-x'\Vert_{2}\leq\varepsilon$. The covering number $N_{\varepsilon}(\mathcal{X})$
is defined as the minimum cardinality of an $\varepsilon$-net
of $\mathcal{X}$. 

We impose the following conditions.
\begin{itemize}
\item (\textbf{Low-dimensionality}) Fix $\varepsilon=T^{-c_{\varepsilon}}$,
where $c_{\varepsilon}>0$ is a sufficiently large absolute constant.
We say that $\mathcal{X}$ has intrinsic dimension $k>0$ if
\[
\log N_{\varepsilon}(\mathcal{X})\leq C_{\mathsf{cover}}k\log T
\]
for some constant $C_{\mathsf{cover}}>0$. 
\item (\textbf{Bounded support}) Assume that, for some universal constant $c_R>0$,
\[
\sup_{x\in\mathcal{X}}\left\Vert x\right\Vert _{2}\leq R\qquad\text{where}\qquad R\coloneqq T^{c_{R}}.
\]
Thus, the radius of the support is allowed to grow polynomially with the number of sampling steps $T$.
\end{itemize}

This formulation accommodates distributions supported on, or concentrated near, low-dimensional manifolds and is therefore less restrictive than assuming an exact linear low-dimensional structure. As a simple sanity check, suppose that $\mathcal{X}$ is contained in an $r$-dimensional subspace of $\mathbb{R}^d$. A standard volumetric argument; see, for example, \citet[Section 4.2.1]{vershynin2018high}; yields
$$
\log N_{\varepsilon}(\mathcal{X})
\asymp
r\log(R/\varepsilon)
\asymp
r\log T.
$$
Hence, in this setting, the intrinsic dimension $k$ is of the same order as the subspace dimension $r$.
\section{Results}
\label{sec:results}

We now present our main theoretical result. 

To characterize the effect of update coefficients, given a noise schedule $(\alpha_t)_{t=1}^T$, we define a sequence of effective noise variances $(v_t)_{t=1}^T$ recursively by
\begin{align}
v_T := 1-\overline{\alpha}_T \quad \text{and}\quad
\alpha_tv_{t-1} = \Big(1 - \frac{\eta_t}{1-\overline{\alpha}_t}\Big)^2v_t + \sigma_t^2, \quad t = T,\ldots, 2. \label{eq:var-defn}
\end{align}
In words, $v_T$ is initialized at the standard terminal noise variance $1 - \overline{\alpha}_T$, and the remaining variances are determined recursively by the update coefficients $\eta_t$ and $\sigma_t^2$.

To provide some intuition, the quantity $v_t$ can be interpreted as the noise level of an auxiliary forward process that is approximately tracked by the sampling iterates. More precisely, we define an auxiliary random process $(\widehat{X}_t)_{t=1}^T$ by
\begin{align}
\widehat{X}_{t} = \sqrt{\overline{\alpha}_t}X_0 + \sqrt{v_t}Z_t, \label{eq:auxiliary}
\end{align} where $Z_t \overset{\mathsf{i.i.d.}}{\sim} \cN(0,I_d)$ is standard Gaussian noise independent of $X_0$.
We further define the noise deviation
\begin{align}
\delta_t \defn \frac{v_t}{1-\overline{\alpha}_t} -1 \label{eq:delta-defn}.
\end{align}
Thus, $\delta_t$ measures the discrepancy between the effective noise level $v_t$ induced by the update coefficients and the noise level $1 - \overline{\alpha}_t$ in the canonical forward process.

We are now ready to state the main theorem.
\begin{theorem}\label{thm:main}
Consider the diffusion sampler defined in \eqref{eq:DDPM} with noise schedule $(\alpha_t)_{t=1}^T$ and update coefficients $(\eta_t, \sigma_t^2)_{t=1}^T$.
Assume that 
\begin{align}
\eta_t \leq C_\eta(1-{\alpha}_t), \quad \sigma_t^2 \geq C_\sigma(1-{\alpha}_t), \quad t\in[T] \label{eq:coeff-condition}
\end{align}
for any constants $C_\eta, C_\sigma>0$.
Suppose that the noise deviation sequence $(\delta_t)_{t=1}^T$ defined in \eqref{eq:delta-defn} satisfies
\begin{align}\label{eq:delta-condition}
|\delta_t| \leq c \frac{\log^2 T}{T}, \quad t\in[T]
\end{align}
for some sufficiently small universal constant $c>0$.
Then the sampler output $Y_1$ satisfies
\begin{align}
\mathsf{TV}\big(\wh X_1, Y_1\big) \leq C\bigg( \frac{k\log^{4} T}{T} + \varepsilon_{\score} \log^{3/2}T \bigg) \label{eq:main-result}
\end{align}
for some universal constant $C>0$, where $\varepsilon_{\score}$ denotes the average score estimation error
\begin{align} \label{eq:score-error}
	\varepsilon_{\score}^2 \defn \frac{1}{T}\sum_{t = 1}^T \mathbb{E}_{\wh X_t\sim p_{\wh X_t}}\Big[\big\|s_t(\wh X_t) - s^\star_t(\wh X_t)\big\|_2^2\Big].
	\end{align}
\end{theorem}

Theorem~\ref{thm:main} shows that, provided the effective variance $v_t$ induced by the update coefficients $(\eta_t, \sigma_t)$ remains close to the canonical forward process variance $1 - \overline{\alpha}_t$, the sampler output $Y_1$ is close in TV distance to the auxiliary random vector $\wh X_1$. Importantly, the leading term that captures the discretization error depends on the intrinsic dimension $k$ rather than the ambient dimension $d$. This theorem therefore demonstrates that low-dimensional adaptation holds for a broad class of coefficient choices, rather than only for a particular discretization scheme. The second term in \eqref{eq:main-result} captures the effect of imperfect score estimation. 
It is worth noting that the score estimation error $\varepsilon_{\score}$ is defined with respect to the auxiliary forward process $(\wh X_t)_{t=1}^T$, rather than the canonical forward process $(X_t)_{t=1}^T$. This is because the sampling iterates $Y_t$ approximately track the auxiliary process $(\wh X_t)_{t=1}^T$ rather than the canonical process $(X_t)_{t=1}^T$.


To accommodate deterministic samplers such as DDIM for which $\sigma_t$ can be zero, we next establish a counterpart of Theorem~\ref{thm:main}. 
The result shows that the discretization error continues to adapt to the intrinsic dimension even in the absence of injected Gaussian noise.
This extension, however, requires access to the exact score functions. Developing a unified theory that simultaneously captures coefficient robustness and score estimation error for deterministic samplers is left for future work.

\begin{theorem}\label{thm:exact-score}
Consider the diffusion sampler defined in \eqref{eq:DDPM} with noise schedule $(\alpha_t)_{t=1}^T$ and update coefficients $(\eta_t, \sigma_t^2)_{t=1}^T$.
Assume that the score estimates are exact, i.e., $s_t=s_t^{\star}$ for all $t\in[T]$, and  
\begin{align}
\eta_t \leq C_\eta(1-{\alpha}_t),  \quad t\in[T]
\end{align}
for any constant $C_\eta>0$.
Suppose that the noise deviation sequence $(\delta_t)_{t=1}^T$ defined in \eqref{eq:delta-defn} satisfies \eqref{eq:delta-condition} for some sufficiently small universal constant $c>0$.
Then the sampler output $Y_1$ satisfies 
\begin{align}
\mathsf{TV}\big(\wh X_1, Y_1\big) \leq C \frac{k\log^{4} T}{T} \label{eq:main-result-2}
\end{align}
for some universal constant $C>0$.
\end{theorem}

We next discuss several important implications of Theorem~\ref{thm:main}.

\paragraph{Low-dimensional adaptation.}
First, consider the idealized setting where the perfect score functions are available, that is, $s_t = s_t^{\star}$ for $t\geq 1$. Suppose that the update coefficients $(\eta_t,\sigma_t)$ satisfy the condition \eqref{eq:delta-condition}. To guarantee $\mathsf{TV}\big({\wh X_1}, {Y_1}\big) \leq \varepsilon$, it then suffices to take
\begin{align*}
\wt O\bigg(\frac{k}{\veps}\bigg)
\end{align*}
sampling iterations.
Consequently, diffusion models achieve substantial sampling acceleration when the target data distributions has intrinsic dimension $k$ much smaller than the ambient dimension $d$. Notably, the sampler does not require prior knowledge of the underlying low-dimensional structure to obtain this speedup.

The following corollary verifies that the condition on $\delta_t$ is satisfied by several important diffusion samplers used in practice.
\begin{corollary}\label{cor:main}
The sampling procedures summarized in \eqref{eq:coeff-choices} satisfy the conditions \eqref{eq:coeff-condition} and \eqref{eq:delta-condition}. Therefore, the convergence bound \eqref{eq:main-result} in Theorem~\ref{thm:main} holds for all these samplers.
\end{corollary}

Corollary~\ref{cor:main} shows that the convergence guarantee $\wt O(k/T)$ in Theorem~\ref{thm:main} achieves the state-of-the-art scaling established in prior works \citep{li2024d,liang2025low}, while applying to a substantially broader range of samplers with various coefficient choices. This demonstrates that adaptation to low-dimensional structure is a robust property of diffusion samplers and does not rely on a narrowly prescribed choice of update coefficients.


\paragraph{Relating the auxiliary marginal to the canonical marginal.}
Prior convergence analysis has primarily focused on $\mathsf{TV}({{X}_1},{Y_1})$ as a measure of how accurately the sampler output approximates the target distribution \citep{li2024d,liang2025low}, while Theorem~\ref{thm:main} controls $\mathsf{TV}({\widehat{X}_1},{Y_1})$. The two reference distributions $X_1$ and $\widehat{X}_1$ differ only in their noise variances. Indeed,
$$
X_1 \disteq \sqrt{\alpha_1}X_0 + \sqrt{1-\alpha_1}Z_1 \quad\text{and}\quad \widehat{X}_1 \disteq \sqrt{\alpha_1}X_0 + \sqrt{v_1}Z_1 = \sqrt{\alpha_1}X_0 + \sqrt{(1+\delta_1)(1-\alpha_1)}Z_1,
$$
where $Z_1\sim N(0,I_d)$ is independent Gaussian noise.
Thus, relative to $X_1$, the Gaussian noise variance in $\widehat{X}_1$ is modified by a factor of $1+\delta_1$. 
%
In particular, for an arbitrary noise schedule $(\alpha_t^{})_{t=1}^T$, there exists a choice of update coefficients $(\eta_t^{},\sigma_t^{})_{t=1}^T$ for which the corresponding noise level deviations $\delta_t^{} $ defined in \eqref{eq:delta-defn} satisfy $\delta_t^{} = 0$ for $1\le t\le T$ \citep{li2024d,liang2025low}. As a result, $X_1 \disteq \wh X_1$ and the corresponding sampler output $Y_1^{}$ satisfies
   \begin{align}\label{eq:cor-optimal}
\mathsf{TV}\big({X_1},{Y_1^{}}\big)\le C\left(\frac{k\log^4T}{T}+\varepsilon_{\mathsf{sc}}\log^{3/2}T\right).
\end{align} 
%

More importantly, exact variance matching does not require replacing a practical sampler by this special choice of update coefficients. For a given coefficient rule, even when the induced deviations $(\delta_t)_{t=1}^T$ are nonzero, one can instead modify the noise schedule $(\alpha_t)_{t=1}^T$ while preserving the functional dependence of $(\eta_t,\sigma_t)$ on the noise schedule $\alpha_t$. The modified noise schedule can be chosen so that the resulting auxiliary marginal has the same signal-to-noise ratio as the original canonical marginal. After a scalar normalization, the two marginals therefore coincide in distribution. This observation is formalized in the following corollary.
\begin{corollary}\label{cor:alpha2}
Consider an arbitrary noise schedule $(\alpha_t^{})_{t=1}^T$ and any choice of update coefficients $(\eta_t,\sigma_t)_{t=1}^T$ satisfying $\eta_t = O(1-\alpha^{}_t)$ and
\begin{align}\label{eq:condition-sigma}
\sigma_t^2 = \alpha^{}_t-\overline{\alpha}^{}_t - \Big(1 - \frac{\eta_t}{1-\overline{\alpha}^{}_t}\Big)^2(1-\overline{\alpha}^{}_t)+O\Big(\frac{(1-\alpha^{}_t)^2}{1-\overline{\alpha}^{}_t}\Big).
\end{align}
These conditions are satisfied by all the sampling procedures summarized in \eqref{eq:coeff-choices}.
Then there exists a noise schedule $(\alpha_t^{\prime})_{t=1}^{T}$ such that the output $Y_1^{\prime}$ associated with the noise schedule $(\alpha_t^{\prime})_{t=1}^{T}$ and update coefficients $(\eta_t,\sigma_t)_{t=1}^T$, after the normalization
$$
\widehat{Y}_1 = C_{\mathsf{rescale}} Y_1^{\prime}
$$
for some constant $C_{\mathsf{rescale}}$ depending on only $\alpha_1^{\prime}$ and $\delta_1^{\prime}$,
satisfies
\begin{align}\label{eq:cor-tv}
\mathsf{TV}\big({X_1},{\widehat{Y}_1}\big)\le C\left(\frac{k\log^4T}{T}+\varepsilon_{\mathsf{sc}}\log^{3/2}T\right),
\end{align}
where $\delta_t^{\prime}$ is the noise level deviation induced by the noise schedule $\alpha_t^{\prime}$.

\end{corollary}

Corollary~\ref{cor:alpha2} combines two observations. First, Theorem~\ref{thm:main} guarantees low-dimensional convergence of $Y_1^\prime$ to the auxiliary marginal $\widehat{X}_1$. Second, the modified noise schedule matches the signal-to-noise ratio of $\widehat{X}_1$ after some normalization with that of the original canonical marginal $X_1$, so that the normalized auxiliary marginal is exactly equal in distribution to $X_1$. Since TV distance is invariant under a common invertible scaling, these two facts together yield the low dimension-dependent convergence guarantee of $\TV(X_1,\widehat{Y}_1)$ in \eqref{eq:cor-tv}. Importantly, this does not require the coefficient-induced deviations $\delta_t$ to vanish or alter the functional relationship between the update coefficients and the noise schedule, and thus covers a broad range of practical samplers.


\paragraph{Comparison with existing lower bounds.}
We conclude this section by comparing our result with existing lower bounds on one-step discretization error, namely, the sampling error incurred when exact score functions are available after a single update step.

For a DDPM sampler with perfect scores, define the one-step update map $\Phi_t^\star:\mathbb{R}^d\times\mathbb{R}^d\to\mathbb{R}^d$ by
\begin{align*}
\Phi_t^\star(x,z) \defn \frac1{\sqrt{\alpha_t}}(x + \eta_ts_t^\star(x) + \sigma_t z).
\end{align*}
The DDPM update rule \eqref{eq:DDPM} can be expressed as $Y_{t-1} = \Phi_t^\star(Y_t,Z_t)$, where $Z_t\sim \cN(0,I_d)$ is independent of $Y_t$.
Starting from the forward marginal $X_t$, prior work \citep{liang2025low} establish the following lower bound on the one-step error:
    \begin{align*}
        \TV\big(\Phi(X_{t},Z_t), X_{t-1}\big) 
        \gtrsim \sqrt{d}\left|\frac{1-\overline{\alpha}_t}{\alpha_t-\overline{\alpha}_t}\left(1-\frac{\eta_t}{1-\overline{\alpha}_t}\right)^2+\frac{\sigma_t^2}{\alpha_t-\overline{\alpha}_t}-1\right|,
    \end{align*}
    where $Z_t\sim \cN(0,I_d)$ is standard Gaussian noise independent of $X_t$.
This lower bound suggests that the TV distance between the sampling iterate and the canonical forward process may scale with the ambient dimension $d$. 
In particular, for this one-step error to vanish as $d$ grows, the coefficient-dependent quantity inside the absolute value must be of order $o(d^{-1/2})$. Viewed solely relative to the canonical forward process, low-dimensional adaptation may therefore appear to fail for coefficient choices that do not satisfy this stringent matching condition.

Our analysis reveals a different mechanism. 
Even when the sampling iterate is not close to the canonical forward marginal $X_{t-1}$, it may remain close to the corresponding auxiliary marginal $\widehat{X}_{t-1}$, whose effective variance incorporates the discrepancy induced by the sampler coefficients. Because this variance discrepancy can either be eliminated through an appropriate coefficient choice or absorbed into a modified noise schedule and final rescaling, the auxiliary process perspective allows us to establish intrinsic dimension-adaptive convergence guarantees for a substantially broader class of diffusion samplers.
    
    
\section{Other related works}

\paragraph*{Adaptation to low-dimensional structures: convergence theory.}

A substantial body of work has developed convergence guarantees for diffusion  samplers, including standard DDPM and DDIM \citep{chen2022sampling,lee2023convergence,chen2023improved,li2023towards,chen2023probability,huang2024convergence,benton2023linear,li2024d,li2025dimension} and higher-order accelerated samplers \citep{li2024provable,li2024improved,jiao2024instance,huang2024convergence,huang2024reverse,li2024accelerating,yu2025advancing,li2025faster,gupta2024faster,wu2024stochastic}.
More recently, \citet{li2024adapting} showed that the sampling convergence rate of diffusion samplers can depend only on the intrinsic dimension of the underlying distribution, and this dependence was subsequently sharpened in \citet{potaptchik2024linear,huang2024denoising,li2024d,liang2025low}.
Beyond continuous diffusion models, analogous low-dimensional adaptation results have also been established for discrete diffusion models. \citep{li2025breaking,zhao2026adaptation,cai2026confidence,chen2025optimal,dmitriev2026efficient}.

\paragraph*{Adaptation to low-dimensional structures: statistical theory.}

Complementing this computational perspective, another strand of the statistical efficiency of diffusion models. 
For general distributions, \citep{oko2023diffusion,wibisono2024optimal,zhang2024minimax,cai2025minimax} demonstrate that diffusion samplers achieve minimax optimal rates for distribution learning and sampling.
A further question is whether these rates improve when the underlying distribution exhibits low-dimensional structure.
For distributions supported on low-dimensional linear subspaces or manifolds, recent works show that the required sample complexity depends on the intrinsic dimension rather than the ambient one \citet{chen2023score,azangulov2024convergence,tang2024adaptivity,wu2026diffusion}. Beyond geometric support assumptions, other works establish minimax-optimal sample   complexity guarantees for structured dependence models \citet{fan2025optimal}, independent component models \citet{boffi2025shallow}, and low-rank Gaussian mixture models \citet{wang2024diffusion}.

\section{Experiment}
\label{sec:exp}

\begin{figure}[t]
    \centering
    \includegraphics[width=0.9\linewidth]{\figpath/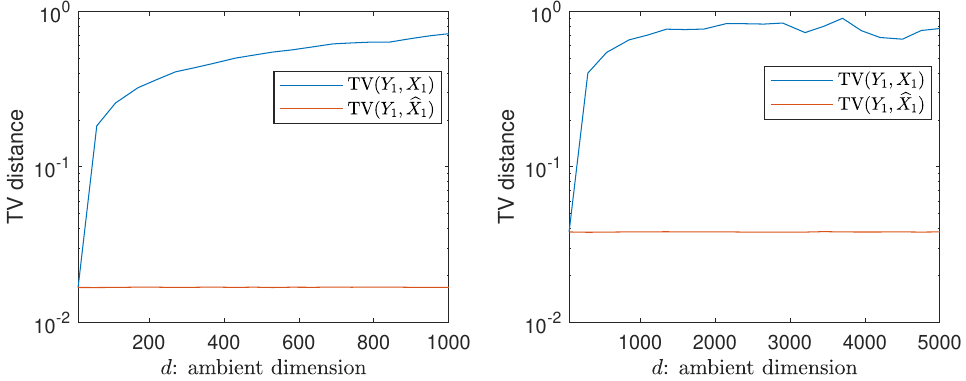}
    \caption{TV distances $\mathsf{TV}({X_1},{Y_1})$ and $\mathsf{TV}({\widehat{X}_1},{Y_1})$ across various ambient dimension $d$.}
    \label{fig:Gauss}
\end{figure}

\begin{figure}[t]
    \centering
    \includegraphics[width=0.9\linewidth]{\figpath/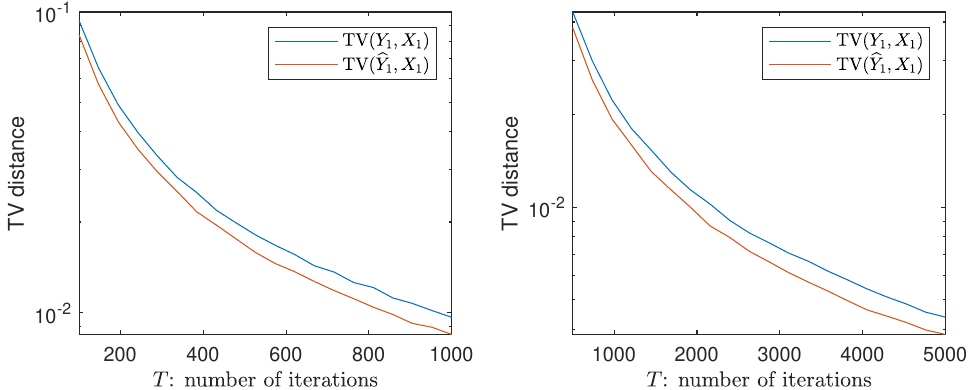}
    \caption{TV distances $\mathsf{TV}({X_1},{{Y}_1^{}})$ and $\mathsf{TV}({{X}_1},{\widehat{Y}_1})$ across various number of iterations $T$.}
    \label{fig:Gauss_cor}
\end{figure}

In this section, we present numerical experiments to validate our theoretical findings.

As it is challenging to compute the TV distance or to examine the influence of intrinsic dimension $k$ and ambient dimension $d$ on real-world datasets, we use a Gaussian distribution to validate our theoretical findings.
Specifically, we consider the case where the target distribution is a $d$-dimensional Gaussian distribution with zero mean and a diagonal covariance matrix for which $k\le d$ entries are equal to $1$ and the remaining entries are zero.
We set $k=10$ or $k=50$, and vary $d$ from $k$ to $5k$.
The total number of iterations is fixed at $T=500$.
After running DDPM with the two parameter choices in \eqref{eq:coeff-choices-ho}, we numerically compute the TV distances $\mathsf{TV}({X_1},{Y_1})$ and $\mathsf{TV}({\widehat{X}_1},{Y_1})$.
The results are shown in Fig. \ref{fig:Gauss}.
They demonstrate that $\mathsf{TV}({X_1},{Y_1})$ is not adaptive to the ambient dimension $d$, whereas $\mathsf{TV}({\widehat{X}_1},{Y_1})$ remains almost unchanged as ambient dimension $d$ increases.
This observation aligns with our theoretical finding in Theorem~\ref{thm:main}.

In addition, to empirically verify Corollary~\ref{cor:alpha2}, we compute the TV distances $\mathsf{TV}(X_1,Y_1^{})$ and $\mathsf{TV}(X_1,\widehat{Y}_1)$ for varying numbers of iterations $T$. Here, $Y_1^{}$ is generated using the update coefficients $(\eta_t^{\star},\sigma_t^{\star})$, while $\widehat{Y}_1$ is obtained using $\eta_t^{\prime} = (\sigma_t^{\prime})^2 = 1-\alpha_t^{\prime}$.
The modified noise schedule $\{\alpha_t^{\prime}\}_{t=1}^T$ is computed numerically using the bisection method.
The results are presented in Fig.~\ref{fig:Gauss_cor}.
As shown in the figure,
the TV distances $\mathsf{TV}(X_1,Y_1^{})$ and $\mathsf{TV}(X_1,\widehat{Y}_1)$ are comparable across different values of $T$, and both decrease as $T$ increases. These observations are consistent with the conclusion of Corollary~\ref{cor:alpha2}.

\section{Discussion}
In this work, we have explored the robustness of adaptation to low-dimensional data structure in diffusion samplers with respect to coefficient choices. We have shown that a broad class of samplers with various coefficient choices can achieve efficient sampling under low-dimensional structure, thereby providing a theoretical justification for the empirical robustness of diffusion samplers in practice.

Looking ahead, several interesting directions remain for future research. First, it would be worthwhile to investigate whether the condition on the effective noise level $v_t$ can be further relaxed, thereby extending the theory to an even broader class of diffusion samplers.
Second, a more refined analysis could characterize the precise influence of the coefficient choices, beyond the order-wise guarantees established in this work.
Third, the current sampling error bound involves a score estimation error evaluated under the auxiliary forward process rather than the canonical forward process used for score training. It would therefore be valuable to determine whether, and to what extent, this distribution shift introduces additional sampling error, and whether it can be controlled under standard score estimation guarantees.

\section*{Acknowledgements}

C.\ Cai is supported in part by the NSF grant DMS-2515333.
G.\ Li is supported in part by the Chinese University of Hong Kong Direct Grant for Research and the Hong Kong Research Grants Council ECS 24305724 and GRF 14307525.

\bibliographystyle{apalike}
\bibliography{bibfileDF}

\appendix
\section{Analysis}
\label{sec:analysis}


\subsection{Preliminaries}

Recall the score function can be expressed as
\begin{align*}
s_{t}^{\star}(x)=-\frac{1}{1-\overline{\alpha}_{t}}\int_{x_{0}}p_{X_{0}|X_{t}}(x_{0}\mymid x)\big(x-\sqrt{\overline{\alpha}_{t}}x_{0}\big)\mathrm{d}x_{0}.
\end{align*}
We define
\begin{align*}
J_{t}(x) &:= \frac{1}{1-\overline{\alpha}_{t}}\bigg\{\Big(\int_{x_{0}}p_{X_{0}|X_{t}}(x_{0}\mymid x)\big(x-\sqrt{\overline{\alpha}_{t}}x_{0}\big)\mathrm{d}x_{0}\Big)\Big(\int_{x_{0}}p_{X_{0}|X_{t}}(x_{0}\mymid x)\big(x-\sqrt{\overline{\alpha}_{t}}x_{0}\big)\mathrm{d}x_{0}\Big)^{\top}\\
 & \qquad\qquad\qquad\quad-\int_{x_{0}}p_{X_{0}|X_{t}}(x_{0}\mymid x)\big(x-\sqrt{\overline{\alpha}_{t}}x_{0}\big)\big(x-\sqrt{\overline{\alpha}_{t}}x_{0}\big)^{\top}\mathrm{d}x_{0}\bigg\}.
\end{align*}
Then the Jacobian matrix of the score function is given by
\begin{align*}
\frac{\partial s_{t}^{\star}(x)}{\partial x} = -\frac{1}{1-\overline{\alpha}_{t}}\big(I + J_{t}(x)\big).
\end{align*}

Next, we record several useful properties of the noise schedule $(\alpha_t)_{t\in[T]}$ defined in \eqref{eq:learning-rate} in Lemma~\ref{lemma:step-size}.
The proof can be found in \citet{li2024provable}.
\begin{lemma}
\label{lemma:step-size}
The learning rates $(\alpha_t)_{t\in[T]}$ specified in \eqref{eq:learning-rate} satisfy the following bounds for all $t=2,\dots,T$:
\begin{subequations}
	\begin{align}
	1-\alpha_t &\leq C_1 \frac{\log T}{T}\label{eq:alpha-t-lb}; \\
	\frac{1-\alpha_t}{1-\ol\alpha_t} &\leq C_1\frac{\log T}{T} \label{eq:1-alpha-1-olalpha}; \\
	\frac{1-\ol\alpha_t}{1-\ol\alpha_{t-1}} &\leq 1+ 2C_1\frac{\log T}{T} \label{eq:1-olalpha-O1}; 
	\end{align}
	where $C_0,C_1$ are defined in \eqref{eq:learning-rate}. In addition, $\alpha_1$ satisfies
	\begin{align}
		1-\alpha_1 & \leq \frac{1}{T^{C_1/4}}\label{eq:alpha-1-lb}.
	\end{align}
\end{subequations}
\end{lemma}

In addition, to facilitate the main analysis, we introduce several auxiliary processes. 
These processes are used solely for the analysis and do not appear in the sampling algorithm.

\paragraph{Auxiliary sequences.}
To begin with, we define an auxiliary reverse process $(Y_t^\star)_{t=0}^T$ that uses the true score functions $\{s_{t}^{\star}(\cdot)\}_{t=1}^T$: 
\begin{equation}
Y_{T}^{\star}\sim\mathcal{N}(0,I_{d}),\quad Y_{t-1}^{\star}\defn\frac{1}{\sqrt{\alpha_{t}}}\big(Y_{t}^{\star}+\eta_t s_{t}^{\star}(Y_{t}^{\star})+\sigma_t Z_{t}\big);\quad t=T,\dots,1.\label{eq:Y-star}
\end{equation}
where $Z_t\overset{\mathrm{i.i.d.}}{\sim}\cN(0,I_d)$ is a sequence of standard Gaussian random vectors independent of $(Y_t^\star)_{t=0}^T$.

We subsequently introduce two auxiliary sequences $(\ol{Y}_{t}^{-})_{t=0}^T$ and $(\ol{Y}_{t})_{t=0}^T$ that capture the discretization error up to some low probability event. Together with $Y_T$, these two sequences form a Markov chain with the transition structure as follows:
\begin{equation}
Y_{T}\to\ol{Y}_{T}^{-}\to\ol{Y}_{T}\to\ol{Y}_{T-1}^{-}\to\ol{Y}_{T-1}\to\cdots\to\ol{Y}_{1}^{-}\to\ol{Y}_{1}\to\ol{Y}_{0}^{-}\to\ol{Y}_{0}.\label{eq:Ybar-chain}
\end{equation}
\begin{itemize}
	\item \emph{Initialization.} For $t=T$, define
\begin{subequations}
\label{eq:Y-bar-init}
	\begin{align}
	\label{eq:Y-bar-init-defn}
		\ol{Y}_{T}^{-}\defn \begin{cases}
		Y_T, & \text{if } Y_T\in\cE_{T}, \\
		\infty, & \text{otherwise}.
		\end{cases}
	\end{align}
	The density of $\ol{Y}_{T}^{-}$ is given by
	\begin{equation}
	p_{\ol{Y}_{T}^{-}}(y)=p_{Y_{T}}(y)\ind\big\{ y\in\mathcal{E}_{T}\big\}+\bP\big\{Y_{T}\notin\mathcal{E}_{T}\big\}\,\delta_{\infty}(y).\label{eq:transition-YTbarminus}
	\end{equation}
\end{subequations}
	\item \emph{Transition from $\ol{Y}_{t}^{-}$ to $\ol{Y}_{t}$.} For $t=T,\dots,0$, we define $\ol{Y}_{t}$ as follows: conditional on $\ol{Y}_{t}^{-}=y$,
\begin{subequations}
	\label{eq:ol-Y-defn}
	\begin{align}
	\ol{Y}_{t}\defn \begin{cases}
	y, & \text{with prob. }  \ol{p}_t(y)/p_{\ol{Y}_{t}^{-}}(y)\wedge 1,\\
	\infty, & \text{with prob. } 1- \big\{ \ol{p}_t(y)/p_{\ol{Y}_{t}^{-}}(y)\wedge 1\big\},
	\end{cases}
	\end{align}  
	where we denote
	\begin{align}
		\ol{p}_t \defn p_{\wh X_{t}},\quad \forall t\geq 0. \label{eq:ol-p-t-defn}
	\end{align}
	The conditional density of $\ol{Y}_{t}$ given $\ol{Y}_{t}^{-}=y$ obeys
	\begin{equation}
	p_{\ol{Y}_{t}\mid \ol{Y}_{t}^{-}}(x \mid y)=\big\{ \ol{p}_t(y)/p_{\ol{Y}_{t}^{-}}(y)\wedge 1\big\}\delta_{y}(x)+\big(1-\big\{ \ol{p}_t(y)/p_{\ol{Y}_{t}^{-}}(y)\wedge1\big\}\big)\delta_{\infty}(x).\label{eq:transition-Ytbarminus-Ytbar}
	\end{equation}
	\end{subequations}
	We make a critical implication of the above construction: for any $t\geq 0$, the density of $\ol Y_t$ satisfies
\begin{align}
\label{eq:ol-Y_t-density}
	p_{\ol Y_t}(y) 
	= \big\{ \ol{p}_t(y)/p_{\ol{Y}_{t}^{-}}(y)\wedge 1\big\} p_{\ol{Y}_{t}^{-}}(y)
	= p_{X_{t}}(y) \wedge p_{\ol Y_t^-}(y),\quad \forall  y\in\bR^d.
\end{align}
\item \emph{Transition from $\ol{Y}_{t}$ to $\ol{Y}_{t-1}^{-}$.} For each $t=T,\dots, 1$, we first generate an intermediate random variable $\widetilde{Y}_{t-1}$ as follows:
\begin{subequations}
\label{eq:ol-Y-mid-defn}
	\begin{align}
	\widetilde{Y}_{t-1}\coloneqq\frac{1}{\sqrt{\alpha_{t}}}\big(\ol{Y}_{t}+\eta_t s_{t}^{\star}(\ol{Y}_{t})+\sigma_t W_{t}\big),
	\end{align}
	where $W_{t}\overset{\mathrm{i.i.d.}}{\sim}\mathcal{N}(0,I_{d}),t\geq 1$ is a sequence of standard Gaussian random vectors independent of $(Z_t)_{t=1}^T$, and then define
	\begin{align}
	\ol{Y}_{t-1}^{-}\defn
	\begin{cases}
	\widetilde{Y}_{t-1}, & \text{if }\ol{Y}_{t}\in\mathcal{E}_{t}\text{ and }\widetilde{Y}_{t-1}\in\mathcal{E}_{t},\\
	\infty, & \text{otherwise.}
	\end{cases}
	\end{align}
	The density of $\ol{Y}_{t-1}^{-}$ given $\ol{Y}_{t}=y$ obeys that when $y\in\mathcal{E}_{t}$,
	\begin{equation}
	p_{\ol{Y}_{t-1}^{-}\mid \ol{Y}_{t}}(x \mid y)=p_{Y_{t-1}^{\star}\mid Y_{t}^{\star}}(x \mid y)\ind\big\{ x \in\mathcal{E}_{t}\big\}
	+\bP\big\{Y_{t-1}^{\star} \notin\mathcal{E}_{t} \mid Y_t^\star = y\big\} \delta_{\infty}(x),\label{eq:transition-Ytbar-Yt-1barminus}
	\end{equation}
	and when $y\notin\mathcal{E}_{t}$,
	\begin{equation}
	p_{\ol{Y}_{t-1}^{-}\mid \ol{Y}_{t}}(x \mid y)=\delta_{\infty}(x).
	\label{eq:transition-Ytbar-Yt-1barminus-2}
	\end{equation}
	\end{subequations}

\end{itemize}

Additionally, we construct two auxiliary sequences $(\wh{Y}_{t}^{-})_{t=0}^T$ and $(\wh{Y}_{t})_{t=0}^T$ that form the following Markov chain:
\begin{equation}
Y_{T}\to\wh{Y}_{T}^{-}\to\wh{Y}_{T}\to\wh{Y}_{T-1}^{-}\to\wh{Y}_{T-1}\to\cdots\to\wh{Y}_{1}^{-}\to\wh{Y}_{1}\to\wh{Y}_{0}^{-}\to\wh{Y}_{0}.\label{eq:Yhat-chain}
\end{equation}
\begin{itemize}
	\item \emph{Initialization.} For $t=T$, initialize $\wh{Y}_{T}^{-} = \ol{Y}_{T}^{-}$.
	\item \emph{Transition from $\wh{Y}_{t}^{-}$ to $\wh{Y}_{t}$.} For $t=T,\dots,0$, the density of $\wh{Y}_{t}$ given $\wh{Y}_{t}^{-}=y$ satisfies
	\begin{equation}
	p_{\wh{Y}_{t}\mid\wh{Y}_{t}^{-}}(x\mid y)=p_{\ol{Y}_{t}\mid\ol{Y}_{t}^{-}}(x\mid y).\label{eq:transition-Ythatminus-Ytbar}
	\end{equation}
\item \emph{Transition from $\wh{Y}_{t}$ to $\wh{Y}_{t-1}^{-}$.} For $t=T,\dots,1$, 
the density of $\ol{Y}_{t-1}^{-}$ given $\ol{Y}_{t}=y$ obeys: when $y\in\mathcal{E}_{t}$,
\begin{subequations}
\label{eq:hat-Y-mid-defn}
	\begin{equation}
	p_{\wh{Y}_{t-1}^{-}\mid \wh{Y}_{t}}(x\mid y)=p_{Y_{t-1}\mid Y_{t}}(x\mid y)\ind\big\{ x\in\mathcal{E}_{t}\big\}
	+\bP\big\{Y_{t-1} \notin\mathcal{E}_{t}\mid Y_t = y\big\} \, \delta_{\infty}(x),\label{eq:transition-Ythat-Yt-1barminus}
	\end{equation}
	and when $y\notin\mathcal{E}_{t}$,
	\begin{equation}
	p_{\wh{Y}_{t-1}^{-}\mid \wh{Y}_{t}}(x\mid y)=\delta_{\infty}(x).
	\label{eq:transition-Ythat-Yt-1barminus-2}
	\end{equation}
	\end{subequations}
\end{itemize}

The sequences $(\wh{Y}_{t}^{-})_{t=0}^T$ and $(\wh{Y}_{t})_{t=0}^T$ are constructed in the same manner as $(\ol{Y}_{t}^{-})_{t=0}^T$ and $(\ol{Y}_{t})_{t=0}^T$, except that the transition from $\wh{Y}_{t}$ to $\wh{Y}_{t}^{-}$ is based on estimated score functions instead of the true score functions.

Moreover, we note that the density of $\wh Y_t$ satisfies
\begin{align}
\label{eq:transition-fact-2}
	p_{\wh Y_t}(x) \leq p_{Y_t}(x), \quad \forall x\in\bR^d,
\end{align}
and $p_{\wh Y_t}(x) \geq p_{Y_t}(x)$ for $x= \infty.$ 
Indeed, observe that the base case $t=T$ is true because $\wh Y_t \overset{\mathrm d}{=} Y_t$, which results from $\wh{Y}_{T}^{-} = \ol{Y}_{T}^{-}$ and $p_{\wh{Y}_{t}\mid\wh{Y}_{t}^{-}}=p_{\ol{Y}_{t}\mid\ol{Y}_{t}^{-}}$ by \eqref{eq:transition-Ythatminus-Ytbar}. Next, assume that
(\ref{eq:transition-fact-2}) holds for $t+1$. Then for any $x\in\bR^d$, we have
\begin{align*}
p_{\widehat{Y}_{t}}(x) & \overset{\text{(i)}}{=}\big\{ p_{X_{t}}(x)/p_{\overline{Y}_{t}^{-}}(x)\wedge 1\big\} p_{\widehat{Y}_{t}^{-}}(x)\leq p_{\widehat{Y}_{t}^{-}}(x) =\int_{\mathbb{R}^{d}}p_{\widehat{Y}_{t}^{-}\mid \widehat{Y}_{t+1}}(x\mid  y)p_{\widehat{Y}_{t+1}}(y)\diff y \\
& \overset{\text{(ii)}}{\leq}\int_{\mathbb{R}^{d}} p_{Y_t\mid Y_{t+1}}(x\mid  y)p_{Y_{t+1}}(y)\diff y=p_{Y_{t}}(x),
\end{align*}
where (i) applies (\ref{eq:transition-Ythatminus-Ytbar})
and (\ref{eq:transition-Ytbarminus-Ytbar}); (ii) arises from the induction hypothesis and \eqref{eq:transition-Ythat-Yt-1barminus}.

\paragraph{Auxiliary sets.}
Let $\{x_i^\star\}_{i=1}^{N_{\varepsilon}}$ be a $\veps$-net of the support of the data distribution $p_\data$, and $\{\cB_i\}_{i=1}^{N_{\varepsilon}}$ be the corresponding $\veps$-cover of the support such that $x_i^\star \in \cB_i$.
We first define sets
\begin{align*}
\mathcal{I} & \coloneqq\left\{ 1\leq i\leq N_{\varepsilon}:\mathbb{P}(X_{0}\in\mathcal{B}_{i})\geq\exp(-\theta k\log T)\right\} ,\\
\mathcal{G} & \coloneqq\big\{\omega\in\mathbb{R}^{d}:\Vert\omega\Vert_{2}\leq2\sqrt{d}+\sqrt{\theta k\log T},\quad\text{and}\\
 & \qquad\qquad\qquad\vert(x_{i}^{\star}-x_{j}^{\star})^{\top}\omega\vert\leq\sqrt{\theta k\log T}\Vert x_{i}^{\star}-x_{j}^{\star}\Vert_{2}\quad\text{for all}\quad1\leq i,j\leq N_{\varepsilon}\big\}.
\end{align*}
where $\theta>0$ is a sufficiently large absolute constant.
Then for each $t=1,\ldots T$, we define a typical set for $\widehat{X}_t$ as follows 
\begin{equation}
\mathcal{E}_{t,1}\coloneqq\left\{ \sqrt{\overline{\alpha}_{t}}\,x_{0}+\sqrt{v_{t}}\,\omega:x_{0}\in\cup_{i\in\mathcal{I}}\mathcal{B}_{i},\omega\in\mathcal{G}\right\},\label{eq:E-t-1-low-d}
\end{equation}
where
\begin{align*}
\Big|\frac{v_{t}}{1-\overline{\alpha}_{t}} - 1\Big| \le \frac{1}{k\log T}.
\end{align*}
Finally, for any $x_{t}\in\mathcal{E}_{t,1}$ and any $r>0$, define set
\begin{equation}
\mathcal{I}\left(x_{t};r\right)\coloneqq\left\{ 1\leq i\leq N_{\varepsilon}:\overline{\alpha}_{t}\Vert x_{i}^{\star}-x_{i(x_{t})}^{\star}\Vert_{2}^{2}\leq r\cdot k(1-\overline{\alpha}_{t})\log T\right\} .\label{eq:I-defn}
\end{equation}

Next, we present several technical lemmas regarding the above auxiliary sets, which will be used in the proof of Theorem~\ref{thm:main}.

\begin{lemma} \label{lem:cond-low-dim} There exists some universal
constant $C_{1}\gg\theta$ such that
\begin{align*}
\mathbb{P}\left(X_{0}\in\mathcal{B}_{i}\mymid X_{t}=x_{t}\right) & \leq\exp\left(-\frac{\overline{\alpha}_{t}}{16\left(1-\overline{\alpha}_{t}\right)}\Vert x_{i(x_{t})}^{\star}-x_{i}^{\star}\Vert_{2}^{2}\right)\mathbb{P}\left(X_{0}\in\mathcal{B}_{i}\right)
\end{align*}
for any $x_{t}\in\mathcal{E}_{t,1}$ and $i\notin\mathcal{I}(x_{t};C_{1}\theta)$.
\end{lemma}

\begin{lemma} \label{lem:Jt} 
There exists some universal constant
$C_{2}\gg C_{1}$ such that for any $x\in\mathcal{E}_{t,1}$,
\begin{equation}
\|J_{t}(x)\|\le\|J_{t}(x)\|_{\frob}\leq\vert\tr(J_{t}(x))\vert\leq C_{2}\theta k\log T.\label{eq:I-Jt-bound-low-d}
\end{equation}
In addition, recall that $\widehat{X}_{t} \disteq \sqrt{\overline{\alpha}_t}X_0 + \sqrt{v_t}Z$ where $Z \sim \mathcal{N}(0, I_d)$ is independent of $X_0$.
There exists some universal constant $C_{0}>0$ such that 
\begin{equation}
\sum_{t=2}^{T}\frac{1-\alpha_{t}}{1-\overline{\alpha}_{t}}\int_{x_{t}}\Vert J_{t}(x_{t})\Vert_{\frob}^{2}\,p_{\widehat{X}_{t}}(x_{t})\diff x_{t}\leq C_{0}k\log T.\label{eq:Jt_sum-low-d}
\end{equation}
 \end{lemma}

\subsection{Proof of Theorems~\ref{thm:main} and \ref{thm:exact-score}}

By the triangle inequality, we can bound the TV distance between $p_{Y_1}$ and $p_{\wh X_1}$ by
\begin{align}
	\mathsf{TV}\big(p_{Y_1}, p_{\wh X_1}\big)
	& \leq \TV(p_{\ol Y_1},\,p_{\wh X_1}) + \TV \big(p_{\ol Y_1},\,p_{Y_1}\big). \label{eq:est-error-temp}
\end{align}
The first term represents the discretization error, up to a low probability event, since $\ol Y_t$ is defined using the true scores. The second term accounts for the error due to score estimation.

In what follows, we bound these two terms separately.

To bound $\TV(p_{\wh X_1},\,p_{\ol Y_1})$, let us first define function $\Delta_{t}(x):\bR^d\rightarrow \bR$, where for each $t=1,\dots,T$:
\begin{equation}
\label{eq:Delta-defn}
\Delta_{t}(x)\coloneqq p_{X_{t}}(x)-p_{\ol{Y}_{t}}(x), \quad \forall x\in\bR^d.
\end{equation}
We know from \eqref{eq:ol-Y_t-density} that $\Delta_t(x)\geq 0$ for all $t\geq 0$ and $x\in\bR^d$. According to the formula for the TV distance $\TV(p,q) = \int_{x\colon p(x)>q(x)}\big(p(x)-q(x)\big)\diff x$, one has
\begin{align}
	\label{eq:TV-ub}
	\TV\big(p_{\wh X_1},\,p_{\ol Y_1}\big) &= \int_{\bR^d\cup\{\infty\}} \big(p_{\wh X_1}(x)-p_{\ol Y_1}(x)\big) \ind\big\{ p_{\wh X_1}(x) > p_{\ol Y_1}(x) \big\} \diff x = \int_{\bR^d}\Delta_1(x)\diff x.
\end{align}

As a result, it suffices to bound $\int \Delta_1(x)\diff x$. To this end, we establish the following lemma, which provides a recursive bound on $\int \Delta_t(x)\diff x$ in terms of $\int \Delta_{t-1}(x)\diff x$.

 \begin{lemma}\label{lem:Delta_t_recursion}
 There exists some universal constant $C_{3}>0$ such that for any $t=1,\ldots,T$,
 \begin{align}
 \int \Delta_{t-1}(x) \diff x \leq \int \Delta_{t}(x) \diff x + C_4 \left(\frac{1-\alpha_t}{1-\ol\alpha_t}\right)^2 \int_{\cE_{t,1}} \big(\big|\tr\big(J_t(x_t)\big)\big| + \|J_t(x_t)\|_{\mathrm{F}}^2\big) p_{\widehat{X}_t}(x_t) \diff x_t + T^{-3}. \label{eq:Delta_t_recursion}
 \end{align}
 In addition, one has $\int \Delta_{T}(x) \diff x \leq T^{-4}$.
 \end{lemma}

 Applying Lemma~\ref{lem:Delta_t_recursion} recursively yields
 \begin{align}
\int \Delta_1(x)\diff x
&\le \int \Delta_T(x)\diff x + T^{-2} + \sum_{t=2}^{T} 
C_4
\left(
\frac{1-\alpha_t}{1-\ol{\alpha}_t}
\right)^2
\int_{x_t\in\mathcal{E}_{t,1}}
\Bigl(
\big|\tr\big(J_t(x_t)\big)\big|+\|J_t(x_t)\|_\frob^2
\Bigr)
p_{\wh X_t}(x_t)\diff x_t \nonumber\\
&\numpf{i}{\le}
8c_1C_4
\frac{\log T}{T}
\sum_{t=2}^{T}
\frac{1-\alpha_t}{1-\ol{\alpha}_t}
\int_{x_t\in\mathcal{E}_{t,1}}
\|J_t(x_t)\|_\frob^2 \,
p_{\wh X_t}(x_t)\diff x_t
+
64c_1^2C_4
\frac{\theta k\log^3 T}{T}
+
T^{-3}
\nonumber\\
&\numpf{ii}{\le}
8c_1C_4C_0
\frac{k\log^2 T}{T}
+
64c_1^2C_4
\frac{\theta k\log^3 T}{T}
+
T^{-3}
\numpf{iii}{\le}
C_5
\frac{k\log^3 T}{T}.
\end{align}
Here, (i), (ii) applies Lemma~\ref{lem:Jt}, and (iii) holds as long as $C_5$ is chosen sufficiently large.
Consequently, we can bound the TV distance between $\wh X_1$ and $\ol Y_1$ as
\begin{align}
\TV(p_{\wh X_1},p_{\ol Y_1})
&=
\int_{p_{\wh X_1}(x)>p_{\ol Y_1}(x)}
\bigl(
p_{\wh X_1}(x)-p_{\ol Y_1}(x)
\bigr)\diff x \notag \\
&=
\int \Delta_1(x)\diff x
\le
C_5\frac{k\log^3 T}{T}. \label{eq:TV-X-olY}
\end{align}

Next, we turn to the term $\TV\big(p_{\ol Y_1},p_{Y_1}\big)$. 
By standard calculations, we have
\begin{align}
\TV\big(p_{\ol{Y}_1},p_{Y_1}\big) & =\int_{\mathbb{R}^{d}}\big(p_{\ol{Y}_1}(x)-p_{Y_1}(x)\big)\ind\big\{ p_{\ol{Y}_1}(x)>p_{Y_1}(x)\big\}\diff x+\mathbb{P}\big\{\ol{Y}_1=\infty\big\}\nonumber \\
 & \overset{\mathrm{(i)}}{\leq}\int_{\mathbb{R}^{d}}\big(p_{\ol{Y}_1}(x)-p_{\wh{Y}_1}(x)\big)\ind\big\{ p_{\ol{Y}_1}(x)>p_{\wh{Y}_1}(x)\big\}\diff x+\mathbb{P}\big\{\ol{Y}_1=\infty\big\}\nonumber \\
 & \overset{\mathrm{(ii)}}{\leq}\TV\big(p_{\ol{Y}_1},p_{\wh{Y}_1}\big)+\TV\big(p_{X_1},p_{\ol{Y}_1}\big) \notag \\
 & \overset{\mathrm{(iii)}}{\leq}\sqrt{\KL\big(p_{\ol{Y}_1}\parallel p_{\wh{Y}_1}\big)}+O\bigg(\frac{k\log^3 T}{T}\bigg).\label{eq:proof-main-6}
\end{align}
where (i) arises from \eqref{eq:transition-fact-2} that $p_{Y_1}(x)\geq p_{\wh Y_1}(x)$ for any $x\in\bR^d$; (ii) uses $\bP\{\ol{Y}_1=\infty\}\leq\TV(p_{\wh X_1},p_{\ol{Y}_1})$ since $\wh X_1 \in \mathbb{R}^{d}$; (iii) applies Pinsker's inequality and \eqref{eq:TV-X-olY}.

To bound the right hand side of \eqref{eq:proof-main-6}, it is sufficient to control $\KL\big(p_{\ol{Y}_1}\parallel p_{\wh{Y}_1}\big)$. 
To this end, observe that
\begin{align}
\KL\big(p_{\ol{Y}_1}\parallel p_{\wh{Y}_1}\big) 
 & \overset{\mathrm{(i)}}{\leq}\KL\big(p_{\ol{Y}_{T}^{-},\ol{Y}_{T},\dots,\ol{Y}_1^{-},\ol{Y}_1}\parallel p_{\wh{Y}_{T}^{-},\wh{Y}_{T},\dots,\wh{Y}_1^{-},\wh{Y}_1}\big)\nonumber \\
 & \overset{\mathrm{(ii)}}{=}\KL\big(p_{\ol{Y}_{T}^{-}}\parallel p_{\wh{Y}_{T}^{-}}\big)+\sum_{t=1}^{T}\mathbb{E}_{x_{t}\sim p_{\ol{Y}_{t}^{-}}}\Big[\KL\big(p_{\ol{Y}_{t}\mid\ol{Y}_{t}^{-}=x_{t}}\parallel p_{\wh{Y}_{t}\mid\wh{Y}_{t}^{-}=x_{t}}\big)\Big] \nonumber \\
 & \quad+\sum_{t=2}^{T}\mathbb{E}_{x_{t}\sim p_{\ol{Y}_{t}}}\Big[\KL\big(p_{\ol{Y}_{t-1}^{-}\mid\ol{Y}_{t}=x_{t}}\parallel p_{\wh{Y}_{t-1}^{-}\mid\wh{Y}_{t}=x_{t}}\big)\Big]\nonumber \\
 & \overset{\mathrm{(iii)}}{=}\sum_{t=2}^{T}\mathbb{E}_{x_{t}\sim p_{\ol{Y}_{t}}}\Big[\KL\big(p_{\ol{Y}_{t-1}^{-}\mid \ol{Y}_{t}=x_{t}}\parallel p_{\wh{Y}_{t-1}^{-}\mid \wh{Y}_{t}=x_{t}}\big)\Big].\label{eq:proof-main-7}
\end{align}
Here, (i) applies the data-processing inequality; (ii) uses the chain rule of KL divergence and the Markov property; (iii) is true since we initialize $\wh{Y}_{T}^{-}=\ol{Y}_{T}^{-}$ and the transition kernels from $\wh{Y}_{t}^{-}$ to $\wh{Y}_{t}$ are the same as those from $\ol{Y}_{t}^{-}$ to $\ol{Y}_{t}$ for all $t\geq 1$.

Observe that $Y_{t-1}^{\star}\mid Y_{t}^{\star}=x_t\sim\mathcal{N}\big(\frac{1}{\sqrt{\alpha_{t}}}\big(x_t+\eta_t s_{t}^{\star}(x_t)\big),\frac{\sigma_t^2}{\alpha_t}I_{d}\big)$ and $Y_{t-1}\mid Y_{t}=x_t\sim\mathcal{N}\big(\frac{1}{\sqrt{\alpha_{t}}}\big(x_t+\eta_t s_{t}(x_t)\big),\frac{\sigma_t^2}{\alpha_t}I_{d}\big)$. 
For any $x_{t}\in \cE_t$, let us denote $p(\cdot) = p_{Y_{t-1}^{\star}\mid Y_{t}^{\star}=x_t}(\cdot)$ and 
$q(\cdot) = p_{Y_{t-1}\mid Y_{t}=x_t}(\cdot)$. Then by the definitions~\eqref{eq:transition-Ythat-Yt-1barminus} and \eqref{eq:transition-Ytbarminus-Ytbar}, we have
\begin{align}
	\KL\big(p_{\ol{Y}_{t-1}^{-}\mid \ol{Y}_{t}=x_{t}}\parallel p_{\wh{Y}_{t-1}^{-}\mid \wh{Y}_{t}=x_{t}}\big) & =\int_{ \cE_t}p(x)\log\frac{p(x)}{q(x)}\diff x+\log\frac{\int_{ \cE_t^\setc}p(x)\diff x}{\int_{ \cE_t^\setc}q(x)\diff x}\int_{ \cE_t^\setc}p(x)\diff x.
\end{align}
Applying \citet[Lemma 6]{li2024d} gives
\begin{align}
\KL\big(p_{\ol{Y}_{t-1}^{-}\mid \ol{Y}_{t}=x_{t}}\parallel p_{\wh{Y}_{t-1}^{-}\mid \wh{Y}_{t}=x_{t}}\big) 
 & \leq\int_{\bR^d}p(x)\log\frac{p(x)}{q(x)}\diff x \notag \\
 & =\KL\big(p_{Y_{t-1}^{\star}\mid Y_{t}^{\star}=x_{t}}\parallel p_{Y_{t-1}\mid Y_{t}=x_{t}}\big)\nonumber \\
 & \overset{\mathrm{(i)}}{=}\frac{\eta_t^2 }{2\sigma^2_t}\big\|  s_{t}(x_{t})-s_{t}^{\star}(x_{t})\big\|_{2}^{2} \notag \\
 & \overset{\mathrm{(ii)}}{\lesssim}\frac{\log T}{T}\big\|  s_{t}(x_{t})-s_{t}^{\star}(x_{t})\big\|_{2}^{2}.\label{eq:proof-main-8}
\end{align}
Here, (i) uses the formula of the KL divergence for two normal distributions; (ii) holds because $\eta_t = O(1-\alpha_t)$, $\sigma_t^2 = \Omega(1-\alpha_t)$, and $1-\alpha_t = O(\log T/T)$.
In addition, for any $x_{t}\in\mathcal{E}_{t}^\setc$, \eqref{eq:transition-Ythat-Yt-1barminus-2} implies that
\begin{equation}
\KL\big(p_{\ol{Y}_{t-1}^{-}|\ol{Y}_{t}=x_{t}}\parallel p_{\wh{Y}_{t-1}^{-}|\wh{Y}_{t}=x_{t}}\big)=0.\label{eq:proof-main-9}
\end{equation}
Combining the above with \eqref{eq:proof-main-7}, we find
\begin{align}
\KL\big(p_{\ol{Y}_1}\parallel p_{\wh{Y}_1}\big) 
& \overset{\mathrm{(i)}}{\leq}\sum_{t=2}^{T}\mathbb{E}_{x_{t}\sim \ol{p}_t}\Big[\KL\big(p_{\ol{Y}_{t-1}^{-}|\ol{Y}_{t}=x_{t}}\parallel p_{\wh{Y}_{t-1}^{-}|\wh{Y}_{t}=x_{t}}\big)\Big] \notag \\
& \overset{\mathrm{(ii)}}{\lesssim} \frac{\log T}{T} \sum_{t=2}^{T}\int_{\bR^d}(1-\overline{\alpha}_t)\| s_{t}(x_{t})-s_{t}^{\star}(x_{t})\|_{2}^{2}\,p_{\wh X_{t}}(x_t)\diff x_t \notag \\
& \overset{\mathrm{(iii)}}{=}\varepsilon_{\score}^2\log T, \label{eq:proof-main-10}
\end{align}
where (i) arises from \eqref{eq:proof-main-9} and \eqref{eq:ol-Y_t-density} that $p_{\ol{Y}_{t}}(x)\leq \ol{p}_t(x)$ for all $x\in\bR^d$; (ii) uses \eqref{eq:proof-main-8}.
Substituting \eqref{eq:proof-main-10} into \eqref{eq:proof-main-6} leads to
\begin{equation}
\TV\big(p_{Y_1},\,p_{\ol{Y}_1}\big)\lesssim \frac{k\log^3 T}{T} + \varepsilon_{\score}\sqrt{\log T}.\label{eq:proof-TV-2-temp}
\end{equation}

Finally, combining \eqref{eq:TV-X-olY} and \eqref{eq:proof-TV-2-temp} yields
\begin{align*}
\TV(p_{\wh X_1},p_{Y_1})
\le
C\left(\frac{k\log^3 T}{T}
+
\varepsilon_{\score}\sqrt{\log T} \right).
\end{align*}

\section{Proof of lemmas and corollaries}
\label{sec:proof-lemma}

\subsection{Proof of Lemma~\ref{lem:cond-low-dim} }
Notice that
\begin{align*}
\mathbb{P}(X_0 \in \mathcal{B}_i \mymid X_t = x_t) &\le \frac{\mathbb{P}(X_0 \in \mathcal{B}_i)\exp(-\frac{\|x_t - \sqrt{\overline{\alpha}_t}x_{i}^{\star}\|^2}{2(1-\overline{\alpha}_t)})}
{\mathbb{P}(X_0 \in \mathcal{B}_{i(x_{t})})\exp(-\frac{\|x_t - \sqrt{\overline{\alpha}_t}x_{i(x_{t})}^{\star}\|^2}{2(1-\overline{\alpha}_t)})} \\
&\le \exp\Big(\frac{\overline{\alpha}_t}{4(1-\overline{\alpha}_t)}\|x_{i}^{\star} - x_{i(x_{t})}^{\star}\|^2 + \theta k\log T\Big)\mathbb{P}(X_0 \in \mathcal{B}_i),
\end{align*}
where the second line holds according to the definition~\eqref{eq:E-t-1-low-d} that
\begin{align*}
\|x_t - \sqrt{\overline{\alpha}_t}x_{i}^{\star}\|^2 - \|x_t - \sqrt{\overline{\alpha}_t}x_{i(x_{t})}^{\star}\|^2
&= \overline{\alpha}_t\|x_{i}^{\star} - x_{i(x_{t})}^{\star}\|^2 - 2\sqrt{\overline{\alpha}_t}(x_{i}^{\star} - x_{i(x_{t})}^{\star})^{\top}(x_t - \sqrt{\overline{\alpha}_t}x_{i(x_{t})}^{\star}) \\
&\ge \frac{\overline{\alpha}_t}{2}\|x_{i}^{\star} - x_{i(x_{t})}^{\star}\|^2.
\end{align*}

\subsection{Proof of Lemma~\ref{lem:Jt} }
Notice that for $x\in\mathcal{E}_{t,1}$,
\begin{align*}
\frac{p_{X_{0}|X_{t}}(x_{0}\mymid x)}{p_{X_{0}|X_{t}}(x_{0}(x)\mymid x)}
&= \frac{p_{X_{0}}(x_{0})}{p_{X_{0}}(x_{0}(x))}\cdot \frac{\exp\big(-\frac{\|x-\sqrt{\overline{\alpha}_{t}}x_{0}\|^{2}}{2(1-\overline{\alpha}_t)}\big)}{\exp\big(-\frac{\|x-\sqrt{\overline{\alpha}_{t}}x_{0}(x)\|^{2}}{2(1-\overline{\alpha}_t)}\big)} \\
&= \frac{p_{X_{0}}(x_{0})}{p_{X_{0}}(x_{0}(x))}\cdot \exp\Big(-\frac{\overline{\alpha}_t\|x_0 - x_0(x)\|^2 - 2\sqrt{\overline{\alpha}_t}(x_0 - x_0(x))^{\top}(x - \sqrt{\overline{\alpha}_t}x_0(x))}{2(1-\overline{\alpha}_t)}\Big) \\
&= c(x_0; x)\frac{p_{X_{0}}(x_{0})}{p_{X_{0}}(x_{0}(x))}\cdot \frac{\exp\big(-\frac{\|x-\sqrt{\overline{\alpha}_{t}}x_{0}\|^{2}}{2v_t}\big)}{\exp\big(-\frac{\|x-\sqrt{\overline{\alpha}_{t}}x_{0}(x)\|^{2}}{2v_t}\big)} 
= c(x_0; x)\frac{p_{X_{0}|\widehat{X}_{t}}(x_{0}\mymid x)}{p_{X_{0}|\widehat{X}_{t}}(x_{0}(x)\mymid x)},
\end{align*}
where $\frac{3}{4} < c(x_0; x) < \frac{4}{3}$ for $\overline{\alpha}_t\|x_0 - x_0(x)\|^2 \lesssim k(1-\overline{\alpha}_t)\log T$.
This implies that
\begin{align*}
p_{X_{0}|X_{t}}(x_{0}\mymid x)
&= \frac{p_{X_{0}|X_{t}}(x_{0}\mymid x)}{\int p_{X_{0}|X_{t}}(x_{0}'\mymid x)\mathrm{d} x_{0}'} \\
&\le \frac{p_{X_{0}|X_{t}}(x_{0}\mymid x)/p_{X_{0}|X_{t}}(x_{0}(x)\mymid x)}{\int_{\overline{\alpha}_t\|x_0' - x_0(x)\|^2 \lesssim k(1-\overline{\alpha}_t)\log T} p_{X_{0}|X_{t}}(x_{0}'\mymid x)/p_{X_{0}|X_{t}}(x_{0}(x)\mymid x)\mathrm{d} x_{0}'} \\
&< \frac{16p_{X_{0}|\widehat{X}_{t}}(x_{0}\mymid x)/p_{X_{0}|\widehat{X}_{t}}(x_{0}(x)\mymid x)}{9\int_{\overline{\alpha}_t\|x_0' - x_0(x)\|^2 \lesssim k(1-\overline{\alpha}_t)\log T} p_{X_{0}|\widehat{X}_{t}}(x_{0}'\mymid x)/p_{X_{0}|\widehat{X}_{t}}(x_{0}(x)\mymid x)\mathrm{d} x_{0}'} \\
&= \frac{16p_{X_{0}|\widehat{X}_{t}}(x_{0}\mymid x)}{9(1-\exp(-\Omega(k\log T)))}
< 2p_{X_{0}|\widehat{X}_{t}}(x_{0}\mymid x).
\end{align*}
Recall that
\begin{align*}
0 &\preceq -J_{t}(x) \preceq \frac{1}{1-\overline{\alpha}_{t}}\bigg\{\int_{x_{0}}p_{X_{0}|X_{t}}(x_{0}\mymid x)\bigg(x-\sqrt{\overline{\alpha}_{t}}x_{0} - \int_{x_{0}}p_{X_{0}|\widehat{X}_{t}}(x_{0}\mymid x)\big(x-\sqrt{\overline{\alpha}_{t}}x_{0}\big)\mathrm{d}x_{0}\bigg) \\
 & \qquad\qquad\qquad\quad\bigg(x-\sqrt{\overline{\alpha}_{t}}x_{0} - \int_{x_{0}}p_{X_{0}|\widehat{X}_{t}}(x_{0}\mymid x)\big(x-\sqrt{\overline{\alpha}_{t}}x_{0}\big)\mathrm{d}x_{0}\bigg)^{\top}\mathrm{d}x_{0}\bigg\} \\
 &\preceq \frac{1}{1-\overline{\alpha}_{t}}\bigg\{\int_{\overline{\alpha}_t\|x_0 - x_0(x)\|^2 \lesssim k(1-\overline{\alpha}_t)\log T}p_{X_{0}|X_{t}}(x_{0}\mymid x)\bigg(x-\sqrt{\overline{\alpha}_{t}}x_{0} - \int_{x_{0}}p_{X_{0}|\widehat{X}_{t}}(x_{0}\mymid x)\big(x-\sqrt{\overline{\alpha}_{t}}x_{0}\big)\mathrm{d}x_{0}\bigg) \\
 & \qquad\qquad\qquad\quad\bigg(x-\sqrt{\overline{\alpha}_{t}}x_{0} - \int_{x_{0}}p_{X_{0}|\widehat{X}_{t}}(x_{0}\mymid x)\big(x-\sqrt{\overline{\alpha}_{t}}x_{0}\big)\mathrm{d}x_{0}\bigg)^{\top}\mathrm{d}x_{0}\bigg\} + \exp(-\Omega(k\log T)).
\end{align*}
As a result, one has
\begin{align*}
-J_{t}(x) &\preceq \frac{2}{1-\overline{\alpha}_{t}}\bigg\{\int_{x_{0}}p_{X_{0}|\widehat{X}_{t}}(x_{0}\mymid x)\bigg(x-\sqrt{\overline{\alpha}_{t}}x_{0} - \int_{x_{0}}p_{X_{0}|\widehat{X}_{t}}(x_{0}\mymid x)\big(x-\sqrt{\overline{\alpha}_{t}}x_{0}\big)\mathrm{d}x_{0}\bigg) \\
 & \qquad\qquad\qquad\quad\bigg(x-\sqrt{\overline{\alpha}_{t}}x_{0} - \int_{x_{0}}p_{X_{0}|\widehat{X}_{t}}(x_{0}\mymid x)\big(x-\sqrt{\overline{\alpha}_{t}}x_{0}\big)\mathrm{d}x_{0}\bigg)^{\top}\mathrm{d}x_{0}\bigg\} + \exp(-\Omega(k\log T)) \\
 &=: -2\widehat{J}_{t}(x) + \exp(-\Omega(k\log T)).
\end{align*}

\subsection{Proof of Lemma~\ref{lem:Delta_t_recursion}}

Observe that
\begin{align}
 & p_{\widehat{X}_{t-1}}(x_{t-1})-\Delta_{t-1}(x_{t-1})+\Delta_{t\to t-1}(x_{t-1})\nonumber \\
 & \ge \int_{x_{t}\in\mathcal{E}_{t}(x_{t-1})}\mathsf{det}\Big(I-\frac{\eta_{t}}{1-\overline{\alpha}_{t}}(I+J_{t}(x_{t}))\Big)^{-1}p_{\widehat{X}_{t}}(x_{t})\Big(\frac{\alpha_{t}}{2\pi\sigma_t^2}\Big)^{d/2}\exp\Big(-\frac{\big\|\sqrt{\alpha_{t}}x_{t-1}-u_{t}\big\|^{2}}{2\sigma_t^2}\Big)\mathrm{d}u_{t}\label{eq:proof-main-1}.
\end{align}

Notice that the density of $\widehat{X}_{t-1}$ admits the following expression:
\begin{align*}
p_{\widehat{X}_{t-1}}(x_{t-1}) &= \int_{u_{t}}\int_{x_{0}} \bigg(2\pi v_t\Big(1 - \frac{\eta_t}{1-\overline{\alpha}_t}\Big)^2\bigg)^{-d/2}p_{X_{0}}(x_{0})\exp\Big(-\frac{\|u_{t}-\sqrt{\overline{\alpha}_{t}}x_{0}\|^{2}}{2(1 - \frac{\eta_t}{1-\overline{\alpha}_t})^2v_t}\Big) \\
&\qquad\qquad \cdot\Big(\frac{\alpha_{t}}{2\pi\sigma_t^2}\Big)^{d/2}\exp\Big(-\frac{\big\|\sqrt{\alpha_{t}}x_{t-1}-u_{t}\big\|^{2}}{2\sigma_t^2}\Big)\mathrm{d}x_{0}\mathrm{d}u_{t}.
\end{align*}
Here we apply the change of variable $u_{t}=x_{t}+\eta_ts_{t}^{\star}(x_{t})$.
In addition, we claim that
\begin{align}
& \mathsf{det}\Big(I-\frac{\eta_{t}}{1-\overline{\alpha}_{t}}(I+J_{t}(x_{t}))\Big)^{-1}\big(2\pi v_t\big)^{-d/2}\int_{x_{0}}p_{X_{0}}(x_{0})\exp\Big(-\frac{\|x_{t}-\sqrt{\overline{\alpha}_{t}}x_{0}\|^{2}}{2 v_t}\Big)\mathrm{d}x_{0} \nonumber\\
& \qquad\qquad= \Big(2\pi v_t\Big(1 - \frac{\eta_t}{1-\overline{\alpha}_t}\Big)^2\Big)^{-d/2}\int_{x_{0}}p_{X_{0}}(x_{0})\exp\Big(-\frac{\|u_{t}-\sqrt{\overline{\alpha}_{t}}x_{0}\|^{2}}{2(1 - \frac{\eta_t}{1-\overline{\alpha}_t})^2v_t}\Big)\mathrm{d}x_{0} \nonumber\\
 & \qquad\qquad\qquad\cdot\exp\bigg(-\xi_{t}(x_{t})+O\Big(\Big(\frac{1-\alpha_{t}}{\alpha_{t}-\overline{\alpha}_{t}}\Big)^{2}\big(k\log T+\|J_{t}(x_{t})\|_{\mathrm{F}}^{2}\big)\Big)\bigg), \label{eq:change}
\end{align}
where $\xi_{t}(x_{t})\le0$ satisfies 
\begin{align}
\int_{x_{t}\in\mathcal{E}_{t,1}}|\xi_{t}(x_{t})|p_{\widehat{X}_{t}}(x_{t})\mathrm{d}x_{t}&\leq C_{3}\Big(\frac{1-\alpha_{t}}{1-\overline{\alpha}_{t}}\Big)^{2}\int_{x_{t}\in\mathcal{E}_{t,1}}\big(k\log T+\|J_{t}(x_{t})\|_{\mathsf{F}}^{2}\big)p_{\widehat{X}_{t}}(x_{t})\mathrm{d}x_{t} + T^{-4}
\end{align}
for some universal constant $C_{3}>0$.
Then the result follows from Lemma~\ref{lem:Jt}.

\paragraph{Proof of Claim~\eqref{eq:change}.}
Notice that
\begin{align*}
 & \frac{\|u_{t}-\sqrt{\overline{\alpha}_{t}}x_{0}\|^{2}}{2(1 - \frac{\eta_t}{1-\overline{\alpha}_t})^2v_t}= \frac{\|x_{t}-\sqrt{\overline{\alpha}_{t}}x_{0}\|_{2}^{2}}{2v_t}+\frac{\frac{\eta_t}{1-\overline{\alpha}_t}(2 - \frac{\eta_t}{1-\overline{\alpha}_t})\|x_{t}-\sqrt{\overline{\alpha}_{t}}x_{0}\|_{2}^{2}}{2(1 - \frac{\eta_t}{1-\overline{\alpha}_t})^2v_t} \\
 & \quad\qquad+\frac{\eta_{t}s_{t}^{\star}(x_{t})^{\top}(x_{t}-\sqrt{\overline{\alpha}_{t}}x_{0})}{(1 - \frac{\eta_t}{1-\overline{\alpha}_t})^2v_t}+\frac{\eta_{t}^{2}\|s_{t}^{\star}(x_{t})\|_{2}^{2}}{2(1 - \frac{\eta_t}{1-\overline{\alpha}_t})^2v_t} \\
 & \quad=\frac{\|x_{t}-\sqrt{\overline{\alpha}_{t}}x_{0}\|^{2}}{2v_t}+\frac{\frac{\eta_t}{1-\overline{\alpha}_t}(2 - \frac{\eta_t}{1-\overline{\alpha}_t})}{2(1 - \frac{\eta_t}{1-\overline{\alpha}_t})^2v_t}\int_{x_{0}}p_{X_{0}|X_{t}}(x_{0}\mymid x_{t})\|x_{t}-\sqrt{\overline{\alpha}_{t}}x_{0}\|_{2}^{2}\mathrm{d}x_{0}\\
 & \quad\qquad-\frac{\frac{\eta_t}{1-\overline{\alpha}_t}(2 - \frac{\eta_t}{1-\overline{\alpha}_t})}{2(1 - \frac{\eta_t}{1-\overline{\alpha}_t})^2v_t}\bigg\|\int_{x_{0}}p_{X_{0}|X_{t}}(x_{0}\mymid x_{t})\big(x_{t}-\sqrt{\overline{\alpha}_{t}}x_{0}\big)\mathrm{d}x_{0}\bigg\|_{2}^{2}+\zeta_{t}(x_{t},x_{0}),
\end{align*}
where we let
\begin{align}
&\zeta_{t}(x_{t},x_{0}) \coloneqq \frac{\frac{\eta_t}{1-\overline{\alpha}_t}(2 - \frac{\eta_t}{1-\overline{\alpha}_t})\big(\|x_{t}-\sqrt{\overline{\alpha}_{t}}x_{0}\|_{2}^{2}-\int_{x_{0}}p_{X_{0}|X_{t}}(x_{0}\mymid x_{t})\|x_{t}-\sqrt{\overline{\alpha}_{t}}x_{0}\|_{2}^{2}\mathrm{d}x_{0}\big)}{2(1 - \frac{\eta_t}{1-\overline{\alpha}_t})^2v_t}\label{eq:zeta-defn-low-d}\\
 & \quad+\frac{\frac{\eta_t}{1-\overline{\alpha}_t}\big[\int_{x_{0}}p_{X_{0}|X_{t}}(x_{0}\mymid x_{t})\big(x_{t}-\sqrt{\overline{\alpha}_{t}}x_{0}\big)\mathrm{d}x_{0}\big]^{\top}\sqrt{\overline{\alpha}_{t}}\big(x_{0}-\int_{x_{0}}p_{X_{0}|X_{t}}(x_{0}\mymid x_{t})x_{0}\mathrm{d}x_{0}\big)}{(1 - \frac{\eta_t}{1-\overline{\alpha}_t})^2v_t}.\nonumber 
\end{align}
Then
\begin{align}
 & \frac{\|u_{t}-\sqrt{\overline{\alpha}_{t}}x_{0}\|^{2}}{2(1 - \frac{\eta_t}{1-\overline{\alpha}_t})^2v_t}
 = \frac{\|x_{t}-\sqrt{\overline{\alpha}_{t}}x_{0}\|^{2}}{2v_t}+\log\det\Big(I-\frac{\eta_{t}}{1-\overline{\alpha}_{t}-\eta_t}J_{t}(x_{t})\Big)\nonumber \\
 & \quad\qquad+\zeta_{t}(x_{t},x_{0})+O\Big(\Big(\frac{1-\alpha_{t}}{\alpha_{t}-\overline{\alpha}_{t}}\Big)^{2}\big(\vert\mathsf{Tr}(J_{t}(x_{t}))\vert+\|J_{t}(x_{t})\|_{\mathrm{F}}^{2}\big)\Big),\label{eq:zeta-decom-low-d}
\end{align}
since
\begin{align*}
\frac{\frac{\eta_t}{1-\overline{\alpha}_t}(2 - \frac{\eta_t}{1-\overline{\alpha}_t})}{2(1 - \frac{\eta_t}{1-\overline{\alpha}_t})^2v_t}
= \frac{\eta_{t}}{1-\overline{\alpha}_{t}-\eta_t} + O\Big(\Big(\frac{1-\alpha_{t}}{\alpha_{t}-\overline{\alpha}_{t}}\Big)^{2}\Big).
\end{align*}

\subsection{Proof of Corollary~\ref{cor:main}}

Through basic algebra, it can be checked that all the choices of coefficients in \eqref{eq:coeff-choices} satisfy 
\begin{align*}
\sigma_t^2 = \alpha_t-\overline{\alpha}_t - \Big(1 - \frac{\eta_t}{1-\overline{\alpha}_t}\Big)^2(1-\overline{\alpha}_t)+O\Big(\frac{(1-\alpha_t)^2}{1-\overline{\alpha}_t}\Big).
\end{align*}
Then we have
\begin{align*}
v_{t-1} = \alpha_t\Big(\frac{1-\overline{\alpha}_{t-1}}{1-\overline{\alpha}_t}\Big)^2v_t + \frac{(1-\alpha_t)(1-\overline{\alpha}_{t-1})}{1-\overline{\alpha}_t} + O\Big(\frac{(1-\alpha_t)^2}{1-\overline{\alpha}_t}\Big),
\end{align*}
which gives
\begin{align*}
v_t = 1-\overline{\alpha}_t + \sum_{i = t+1}^T \frac{\overline{\alpha}_{i}}{\overline{\alpha}_{t}}\Big(\frac{1-\overline{\alpha}_{t}}{1-\overline{\alpha}_i}\Big)^2 \cdot O\Big(\frac{(1-\alpha_{i-1})\log T}{T}\Big)
= (1-\overline{\alpha}_t) \cdot \Big(1 + O\Big(\frac{\log^2 T}{T}\Big)\Big),
\end{align*}
provided that $\frac{1-\alpha_t}{1-\overline{\alpha}_t} \lesssim \frac{\log T}{T}$.
As a result, we conclude
\begin{align*}
|\delta_t| \lesssim \frac{\log^2 T}{T}.
\end{align*}

\subsection{Proof of Corollary~\ref{cor:alpha2} }
\label{sec:proof-cor-alpha2}

    Consider 
    $$
    \eta_t^{\star} \defn 1- \alpha_t, \qquad \sigma_t^{\star} = \sqrt{\frac{\alpha_t-\overline{\alpha}_{t}}{1-\overline{\alpha}_t}(1-\alpha_t)}.
    $$
   Denote $v_t^{\star}$ as the effective noise level in \eqref{eq:delta-defn} induced by schedule $\alpha_t$.
    For $t=T$, we have $v_T^{\star}=1-\overline{\alpha}_T$ and $\delta_T=0$.
    we prove $\delta_t=0$ immediately by induction and verifying that
\begin{align}
    \Big(1 - \frac{\eta_t^{\star}}{1-\overline{\alpha}_t}\Big)^2v_t^{\star} + (\sigma_t^{\star})^2 &= \Big(1 - \frac{1-\alpha_t}{1-\overline{\alpha}_t}\Big)^2(1-\overline{\alpha}_t) + \frac{\alpha_t-\overline{\alpha}_{t}}{1-\overline{\alpha}_t}(1-\alpha_t)\notag\\
    &=\frac{(\alpha_t-\overline{\alpha}_t)^2}{1-\overline{\alpha}_t} + \frac{\alpha_t-\overline{\alpha}_{t}}{1-\overline{\alpha}_t}(1-\alpha_t)\notag\\
    &=\frac{\alpha_t-\overline{\alpha}_t}{1-\overline{\alpha}_t}\left(\alpha_t-\overline{\alpha}_t+1-\alpha_t\right)\notag\\
    &=\alpha_t(1-\overline{\alpha}_{t-1}) = \alpha_tv_{t-1}^{\star}.
\end{align}
By Theorem \ref{thm:main}, we have $\widehat{X}_1 = X_1$ and establish \eqref{eq:cor-optimal}.

Next we intend to prove that there exists a learning rate schedule $\alpha_t^{\prime}$ satisfying 
\begin{align}\label{eq:cor-condition-alpha2}
\frac{1-\overline{\alpha}_t}{\overline{\alpha}_t} = \frac{(1+\delta_t^{\prime})(1-\overline{\alpha}_t^{\prime})}{\overline{\alpha}_t^{\prime}},\qquad 1\le t\le T.
\end{align}
Denote $v_t$ as the effective noise level in \eqref{eq:delta-defn} induced by schedule $\alpha_t^{\prime}$.
Then we have the following equation
\begin{align}\label{eq:cor-Xhat}
\frac{\widehat{X}_1^{\prime}}{\sqrt{\alpha_1^{\prime} + (1+\delta_1^{\prime})(1-\alpha_1^{\prime})}} = X_1 = \sqrt{\alpha_1}X_0 + \sqrt{1-\alpha_1} Z_1,
\end{align}
since
\begin{align*}
\widehat{X}_1^{\prime} &= \sqrt{\alpha_1^{\prime}}X_0 + \sqrt{v_1}Z_1 = \sqrt{\alpha_1^{\prime}}X_0 + \sqrt{(1+\delta_1^{\prime})(1-\alpha_1^{\prime})}Z_1\notag\\
& = \sqrt{\alpha_1^{\prime} + (1+\delta_1^{\prime})(1-\alpha_1^{\prime})}\left(\sqrt{\alpha_1}X_0 + \sqrt{1-\alpha_1} Z_1\right).
\end{align*}
Then we have
\begin{align*}
\mathsf{TV}\left(p_{X_1},p_{\widehat{Y}_1}\right) = \mathsf{TV}\left(p_{\widehat{X}_1^{\prime}/\sqrt{\alpha_1^{\prime} + (1+\delta_1^{\prime})(1-\alpha_1^{\prime})}},p_{\widehat{Y}_1}\right) = \mathsf{TV}(p_{\widehat{X}_1^{\prime}},p_{Y_1^{\prime}}) 
\end{align*}
and by applying Theorem \ref{thm:main}, we complete the proof.

Now it suffices to prove that there exists $\alpha_t^{\prime}$ satisfying \eqref{eq:cor-condition-alpha2}.
We prove this by induction.
Given $\overline{\alpha}_t^{\prime}$ and $v_t$ obeying
\begin{align}\label{eq:proof-cor-2}
v_t = \frac{\overline{\alpha}_t^{\prime}(1-\overline{\alpha}_t)}{\overline{\alpha}_t}.
\end{align}
We need to show the existence of $\overline{\alpha}_t^{\prime}$ (or $\alpha_t^{\prime}$ equivalently) such that the above relation holds also for $t-1$.
Recall that
\begin{align}\label{eq:proof-cor-recurv}
\alpha_t^{\prime}v_{t-1} = \left(1-\frac{\eta_t^{\prime}}{1-\overline{\alpha}_t^{\prime}}\right)^2v_t + (\sigma_t^{\prime})^2.
\end{align}
For fixed $\overline{\alpha}_t^{\prime}$ and $v_t$, we treat $v_{t-1}$ as a continuous function of $\alpha_t^{\prime}$, and examine its values at two specific values of $\alpha_t^{\prime}$ in the following.
\begin{itemize}
\item
When $\overline{\alpha}_{t-1}=\overline{\alpha}_t$, we have $\eta_t^{\prime} = O(1-\alpha_t^{\prime}) = 0$ and 
$$
(\sigma_t^{\prime})^2 = 1 - \overline{\alpha}_t^{\prime} - (1-\overline{\alpha}_t^{\prime}) = 0.
$$
Thus we have
\begin{align*}
    v_{t-1} = v_t = \frac{\overline{\alpha}_t^{\prime}(1-\overline{\alpha}_t)}{\overline{\alpha}_t} = \frac{\overline{\alpha}_{t-1}^{\prime}(1-\overline{\alpha}_t)}{\overline{\alpha}_t} \ge \frac{\overline{\alpha}_{t-1}^{\prime}(1-\overline{\alpha}_{t-1})}{\overline{\alpha}_{t-1}}.
\end{align*}
Recall that $\overline{\alpha}_1$ satisfy
$$
\frac{1-{\alpha}_1}{1-\overline{\alpha}_1} \le \frac{C_1\log T}{T}.
$$
\item
When 
$$
{\alpha}_{t}^{\prime}=1-\frac{C_2\log T}{T}(1-\overline{\alpha}_t^{(t)}),
$$
with $C_2\ge 3C_1$ is a sufficiently large constant, we intend to prove that
\begin{align}\label{eq:proof-cor-4}
v_{t-1} \le\frac{\overline{\alpha}_{t-1}^{\prime}(1-\overline{\alpha}_{t-1})}{\overline{\alpha}_{t-1}}.
\end{align}
Towards this, we insert $\eta_t^{\prime}=O(1-\alpha_t^{\prime})$ and \eqref{eq:condition-sigma} into \eqref{eq:proof-cor-recurv} and obtain
\begin{align}\label{eq:proof-cor-1}
\alpha_t^{\prime}v_{t-1} &= \left(1-\frac{\eta_t^{\prime}}{1-\overline{\alpha}_t^{\prime}}\right)^2v_t + (\sigma_t^{\prime})^2 \notag\\
&= \left(1-\frac{\eta_t^{\prime}}{1-\overline{\alpha}_t^{\prime}}\right)^2(v_t-1+\overline{\alpha}_t^{\prime}) + \alpha_t^{\prime}-\overline{\alpha}_t^{\prime} + O\Big(\frac{(1-\alpha_t^{\prime})^2}{1-\overline{\alpha}_t^{\prime}}\Big)\notag\\
&=v_t-1+\overline{\alpha}_t^{\prime} + \bigg(\left(1-\frac{\eta_t^{\prime}}{1-\overline{\alpha}_t^{\prime}}\right)^2-1\bigg)(v_t-1+\overline{\alpha}_t^{\prime}) + \alpha_t^{\prime}-\overline{\alpha}_t^{\prime}\notag\\
&\qquad\qquad + O\Big(\frac{(1-\alpha_t^{\prime})^2}{1-\overline{\alpha}_t^{\prime}}\Big).
\end{align}
Recall that $\frac{1-\alpha_t^{\prime}}{1-\overline{\alpha}_t^{\prime}} = \frac{C_2\log T}{T}$, we have
\begin{align*}
\frac{(1-\alpha_t^{\prime})^2}{1-\overline{\alpha}_t^{\prime}} &= (1-\overline{\alpha}_t^{\prime})\frac{C_2^2\log^2 T}{T^2},\notag\\
\left|1-\left(1-\frac{\eta_t^{\prime}}{1-\overline{\alpha}_t^{\prime}}\right)^2\right|&\le \frac{2\eta_t^{\prime}}{1-\overline{\alpha}_t^{\prime}} + \left(\frac{\eta_t^{\prime}}{1-\overline{\alpha}_t^{\prime}}\right)^2\notag\\
& = O\left(\frac{1-\alpha_t^{\prime}}{1-\overline{\alpha}_t^{\prime}}\right) + O\left(\left(\frac{1-\alpha_t^{\prime}}{1-\overline{\alpha}_t^{\prime}}\right)^2\right)
=O\left(\frac{\log T}{T}\right),\notag\\
v_t-1+\overline{\alpha}_t^{\prime} &= O\left((1-\overline{\alpha}_t^{\prime})\frac{\log^2T}{T}\right).
\end{align*}
Combining all of these into \eqref{eq:proof-cor-1}, we have
\begin{align}\label{eq:proof-cor-3}
\alpha_t^{\prime}v_{t-1} &=
v_t-1+\overline{\alpha}_t^{\prime} + O\left(\frac{\log T}{T}\right)O\left((1-\overline{\alpha}_t^{\prime})\frac{\log^2T}{T}\right)\notag\\
&\qquad\quad + \alpha_t^{\prime}-\overline{\alpha}_t^{\prime}+ O\left((1-\overline{\alpha}_t^{\prime})\frac{\log^2 T}{T^2}\right)\notag\\
&=v_t-1+\alpha_t^{\prime} + O\left((1-\overline{\alpha}_t^{\prime})\frac{\log^3 T}{T^2}\right).
\end{align}
We can establish \eqref{eq:proof-cor-4} by verifying that 
\begin{align*}
&\alpha_t^{\prime}v_{t-1}  - \alpha_t^{\prime}\frac{\overline{\alpha}_{t-1}^{\prime}(1-\overline{\alpha}_{t-1})}{\overline{\alpha}_{t-1}} \notag\\
&\quad= v_t-1+\alpha_t^{\prime} + O\left((1-\overline{\alpha}_t^{\prime})\frac{\log^3 T}{T^2}\right) - \frac{\overline{\alpha}_{t}^{\prime}(1-\overline{\alpha}_{t-1})}{\overline{\alpha}_{t-1}}\notag\\
&\quad=\frac{\overline{\alpha}_{t}^{\prime}(1-{\alpha}_{t})}{\overline{\alpha}_{t}}-1+\alpha_t^{\prime} + O\left((1-\overline{\alpha}_t^{\prime})\frac{\log^3 T}{T^2}\right)\notag\\
&\quad=(1-\overline{\alpha}_t^{\prime})\left(\frac{\overline{\alpha}_{t}^{\prime}(1-\overline{\alpha}_{t})}{(1-\overline{\alpha}_{t}^{\prime})\overline{\alpha}_{t}}\frac{1-{\alpha}_{t}}{1-\overline{\alpha}_{t}}-\frac{1-\alpha_t^{\prime}}{1-\overline{\alpha}_t^{\prime}}\right)+ O\left((1-\overline{\alpha}_t^{\prime})\frac{\log^3 T}{T^2}\right)\le 0,
\end{align*}
where the last inequality comes from 
\begin{align*}
\frac{\overline{\alpha}_{t}^{\prime}(1-\overline{\alpha}_{t})}{(1-\overline{\alpha}_{t}^{\prime})\overline{\alpha}_{t}}\frac{1-{\alpha}_{t}}{1-\overline{\alpha}_{t}}-\frac{1-\alpha_t^{\prime}}{1-\overline{\alpha}_t^{\prime}} 
&\le \frac{C_1\log T}{T}\frac{\overline{\alpha}_{t}^{\prime}(1-\overline{\alpha}_{t})}{(1-\overline{\alpha}_{t}^{\prime})\overline{\alpha}_{t}}-\frac{C_2\log T}{T}\notag\\
& = \frac{C_1\log T}{T}\frac{v_t}{(1-\overline{\alpha}_{t}^{\prime})}\frac{\overline{\alpha}_{t}^{\prime}(1-\overline{\alpha}_{t})}{v_t\overline{\alpha}_{t}}-\frac{C_2\log T}{T}\notag\\
& = \frac{C_1\log T}{T}(1+\delta_t^{\prime})-\frac{C_2\log T}{T} \le -\frac{C_1\log T}{T}.
\end{align*}
\end{itemize}
By mean value theorem, we conclude that there exists $\alpha_t^{\prime}\in [1-\frac{C_2\log T}{T}(1-\overline{\alpha}_t^{\prime}),1]$, such that \eqref{eq:proof-cor-2} holds for $t-1$.

\end{document}